\documentclass[journal]{IEEEtran}
\usepackage{graphicx}
\usepackage{amsmath}
\usepackage{cite}
\usepackage{times}

\usepackage{array}
\usepackage{rotating}
\usepackage{framed}
\usepackage{multirow}
\usepackage{booktabs}
\usepackage{amsmath}
\usepackage{mathrsfs} 
\usepackage{amssymb}
\usepackage{array}
\usepackage{rotating}
\usepackage{framed}
\usepackage{multirow}
\usepackage{booktabs}
\usepackage{textcomp}
\usepackage{color}

\def\I{\mathbf{I}}
\def\U{\mathbf{U}}
\def\T{T}
\def\E{E}
\def\N{N}

\usepackage{float}
\usepackage{xspace}

\newcommand{\latinphrase}[1]{\textit{#1}}  

\newcommand{\ie}{\latinphrase{i.e.}\xspace}

\newcommand{\eg}{\latinphrase{e.g.}\xspace}

\newcommand{\argmax}{\operatornamewithlimits{argmax}}
\newcommand{\argmin}{\operatornamewithlimits{argmin}}

\begin{document}

\title{Deep Underwater Image Enhancement}

\author{Saeed~Anwar,
        Chongyi~Li,
        Fatih~Porikli
\thanks{This work was supported by the Australian Research Council's Discovery Projects funding scheme (project DP150104645) and the National Natural Science Foundation of China (61771334).}

\thanks{Saeed Anwar and Fatih  Porikli are with Research School of Engineering, Australian National University (ANU), Canberra, ACT 0200, Australia. (e-mail: saeed.anwar@anu.edu.au; fatih.porikli@anu.edu.au).

Chongyi Li is with College of Electronic Information and Automation, Civil Aviation University of China (CAUC), Tianjin 300300, China. (e-mail: lichongyi@tju.edu.cn).

Saeed Anwar and Chongyi Li contribute equally to this work. 

(Corresponding author: Chongyi Li)

}}
\markboth{}%
{Shell \MakeLowercase{\textit{et al.}}: Bare Demo of IEEEtran.cls for Journals}

\maketitle

\begin{abstract}

In an underwater scene, wavelength-dependent light absorption and scattering degrade the visibility of images, causing low contrast and distorted color casts. To address this problem, we propose a convolutional neural network based image enhancement model, \ie, UWCNN, which is trained efficiently using a synthetic underwater image database. Unlike the existing works that require  the parameters of underwater imaging model estimation or impose inflexible frameworks applicable only for specific scenes, our model directly reconstructs the clear latent underwater image by leveraging on an automatic end-to-end and data-driven training mechanism. Compliant with underwater imaging models and optical properties of underwater scenes, we first synthesize ten different marine image databases. Then, we separately train multiple UWCNN models for each underwater image formation type. Experimental results on real-world and synthetic underwater images demonstrate that the presented method generalizes well on different underwater scenes and outperforms the existing methods both qualitatively and quantitatively. Besides, we conduct an ablation study to demonstrate the effect of each component in our network.
\end{abstract}


\section{Introduction}

\begin{figure}[t]
\centering
\includegraphics[width=7cm,height=4cm]{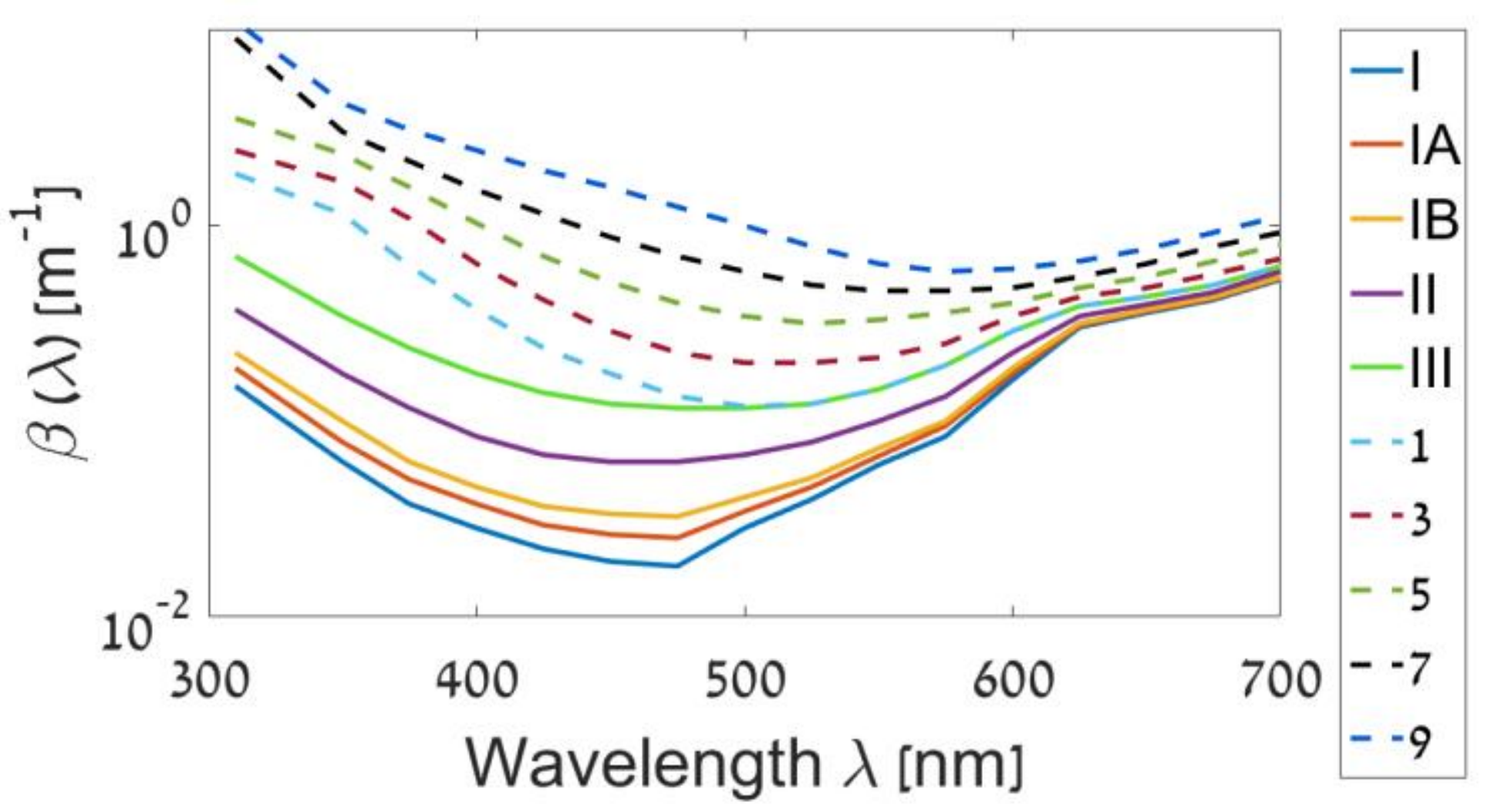}
\caption{Wavelength-dependent light attenuation coefficients $\beta$ of Jerlov water types from \cite{Berman2017bmvc}. Solid lines mark open ocean water types while dashed lines mark coastal water types. The Jerlov water types are I, IA, IB, II and III for open ocean waters, and 1 through 9 for coastal waters. Type-I is the clearest and Type-III is the most turbid open ocean water. Likewise, for coastal waters, Type-1 is clearest and Type-9 is the most turbid \cite{Jaffe1976}.}
\label{fig:Jerlov}
\end{figure}

Acquisition of clear underwater images is of great importance for ocean engineering and ocean research where autonomous and remotely operated underwater vehicles are widely used to explore and interact with marine environments. However, raw underwater images seldom meet the expectations concerning image visual quality.  Naturally, underwater images are degraded by the adverse effects of light absorption and scattering due to particles in the water including micro phytoplankton colored dissolved organic matter and non-algal particles \cite{Mobley41994}. When the light propagates in an underwater scenario, the light received by a camera is mainly composed of three types of light: direct light, forward scattering light, and backscattering light. The direct light suffers from attenuation resulting in information loss of underwater images. The forward scattering light has a negligible contribution to the blurring of the image features. The backscattering light reduces the contrast of underwater images and suppresses fine details and patterns. Additionally, the red light first disappears, followed by the green and blue lights (the wavelengths of the red, green and blue lights are 600nm, 525nm, and 475nm, respectively).  As a result, most underwater images are dominated by a bluish or greenish tone. Figure~\ref{fig:Jerlov} presents a diagram of light attenuation with respect to the wavelength of light.

These absorption and scattering problems hinder the performance of underwater scene understanding and computer vision applications such as aquatic robot inspection \cite{Forest12001} and marine environmental surveillance \cite{Strachan21993}. Therefore, it is necessary to develop effective solutions to improve the visibility, contrast, and color properties of underwater images for a superior visual quality and appeal.

Notable progress has been made to improve the visual quality of underwater images in recent years (see \cite{Schettini62010} for a survey). Existing underwater images sharpness methods can be classified into one of three broad categories: image enhancement methods, image restoration methods, and supplementary-information specific methods.

The image enhancement techniques modify the image pixel values to produce a subjectively and aesthetically pleasing image for specific objectives (\eg, contrast improvement, denoising, and brightness enhancement) without relying on any physical imaging models. In this line of research, \cite{Ancuti82012} presented a fusion-based method to improve the visual quality of underwater images and videos. The method of \cite{Ancuti82012} fuses a contrast improved underwater image and a color corrected underwater image obtained from input. In the process of mulit-scale fusion, four weights are used to determine which pixel is advantaged to appear in the final image. This method improves the global contrast and visibility; however, some regions in the resultant images become over-enhanced or under-enhanced. \cite{Ghani92015, Ghani102015} suggested a Rayleigh-stretched contrast-limited adaptive histogram method which is a modified version of \cite{Iqbal71010}'s method.  This method constrains the number of under-enhanced and over-enhanced regions. Nevertheless, it also tends to increase noise in the resultant images. A hybrid method based on color correction and underwater image dehazing for underwater image enhancement was proposed in \cite{Li2017prl}, which corrects the color casts of underwater image using image color prior and improves the visibility by a modified image dehazing algorithm. This method shows limitations when the image color prior is not available. \cite{Li2018} proposed an underwater image color correction method based on weakly supervised color transfer, which learns a cross domain mapping function between underwater images and air images. This method relaxes the need for paired underwater images for training and allows the underwater images being taken in unknown locations. Recently, \cite{Ancuti2018TIP} modified their previous work \cite{Ancuti82012} in order to reduce the effects of the over-enhancement and over-exposure.

The image restoration methods (\eg, \cite{Carlevaris-Bianco112010}, \cite{Chiang122012}, \cite{Galdran142015}, \cite{Li2015JEI}, \cite{Drews152016}, \cite{Li2016ICIP}, \cite{Li32016}, \cite{Peng2017}, \cite{Li2017}, \cite{Peng2018}) consider the challenge at hand as an inverse problem, and construct physical models of the degradation, and then estimate model parameters.  \cite{Carlevaris-Bianco112010} proposed a prior that exploited the difference in attenuation among the three color channels to predict the medium transmission of underwater scene. As a result, the effects of light scattering in underwater image are removed. However, such a method shows limitations when it is used to process underwater images which have not the strong difference in attenuation among three color channels of an underwater image. \cite{Chiang122012} combined an image dehazing algorithm with a wavelength dependent compensation algorithm to restore underwater image, which can remove the bluish tone of underwater images and the effects of the artificial light. However, this method is limited in processing the underwater images with serious color casts. A Red Channel method presented by \cite{Galdran142015} recovered the lost contrast of the underwater image by restoring the colors associated with short wavelengths. Similarly, \cite{Drews152016} proposed an underwater dark-channel prior called UDCP which modifies the dark channel prior presented in \cite{He132011}. With the proposed UDCP, the medium transmission can be estimated in some cases; however, underwater dark-channel prior does not always hold when there are white objects or artificial light in the underwater scenes. \cite{Li32016} combined an underwater image dehazing algorithm with a contrast enhancement algorithm. This method can yield two enhanced underwater images at the same time. One image with relative genuine color and natural appearance is suitable for display while another image with high  contrast and brightness can be used for unveiling more image details. Recently, \cite{Li2017} proposed a CNN based real-time underwater image color correction model based on synthetic underwater images generated in the weakly supervised learning manner. Nevertheless, this method is only effective for underwater images captured under specific scenes that have to be similar to those of its training data. It is, thus, impractical for real applications. Very recently, \cite{Peng2018} proposed a generalized dark channel prior, which estimates the medium transmission by calculating the difference between the observed intensity and the ambient light. After that, the degraded images are recovered according to an image formation model.

The supplementary-information specific methods such as \cite{Schechner172004}, \cite{Schechner182005}, \cite{Treibitz192009},  and \cite{Sheinin202015} usually take advantage of the additional information obtained from the multiple images captured by polarization filters, stereo images, rough depth of the scene, or specialized hardware devices.

Despite these recent efforts, existing approaches have still the following issues:

\begin{enumerate}

\item The perceived contrast in the latent image is often erratically distributed, remaining ineffectively low in under-enhanced  regions and distractingly high in over-enhanced regions. Moreover, some methods introduce artifacts.

\item Most methods directly employ generic outdoor haze models to predict the underwater imaging model parameters, which is not well-suited for marine scenarios as the nature of imaging and lighting in such environments are different.

\item The specialized sensors and use of multiple images can be prohibitive, expensive and time-consuming, which reduce their applicability.

\item Although deep learning has achieved impressive performance on low-level vision tasks, there is no effective deep model for underwater image enhancement. The main reason is due to lacking of sufficiently labeled training data, which limits the development of deep learning based underwater image enhancement methods.
\end{enumerate}

In contrast, we propose a new underwater image synthesis method, and then design to offer a robust and data-driven solution to solve these issues. We propose a deep convolutional neural network, which is shown to have superior robustness, accuracy, and flexibility for varying water types for marine imaging applications.

\begin{figure}[t]
\begin{center}
\begin{tabular}{c@{ } c}
\includegraphics[width=4cm,height=3.6cm]{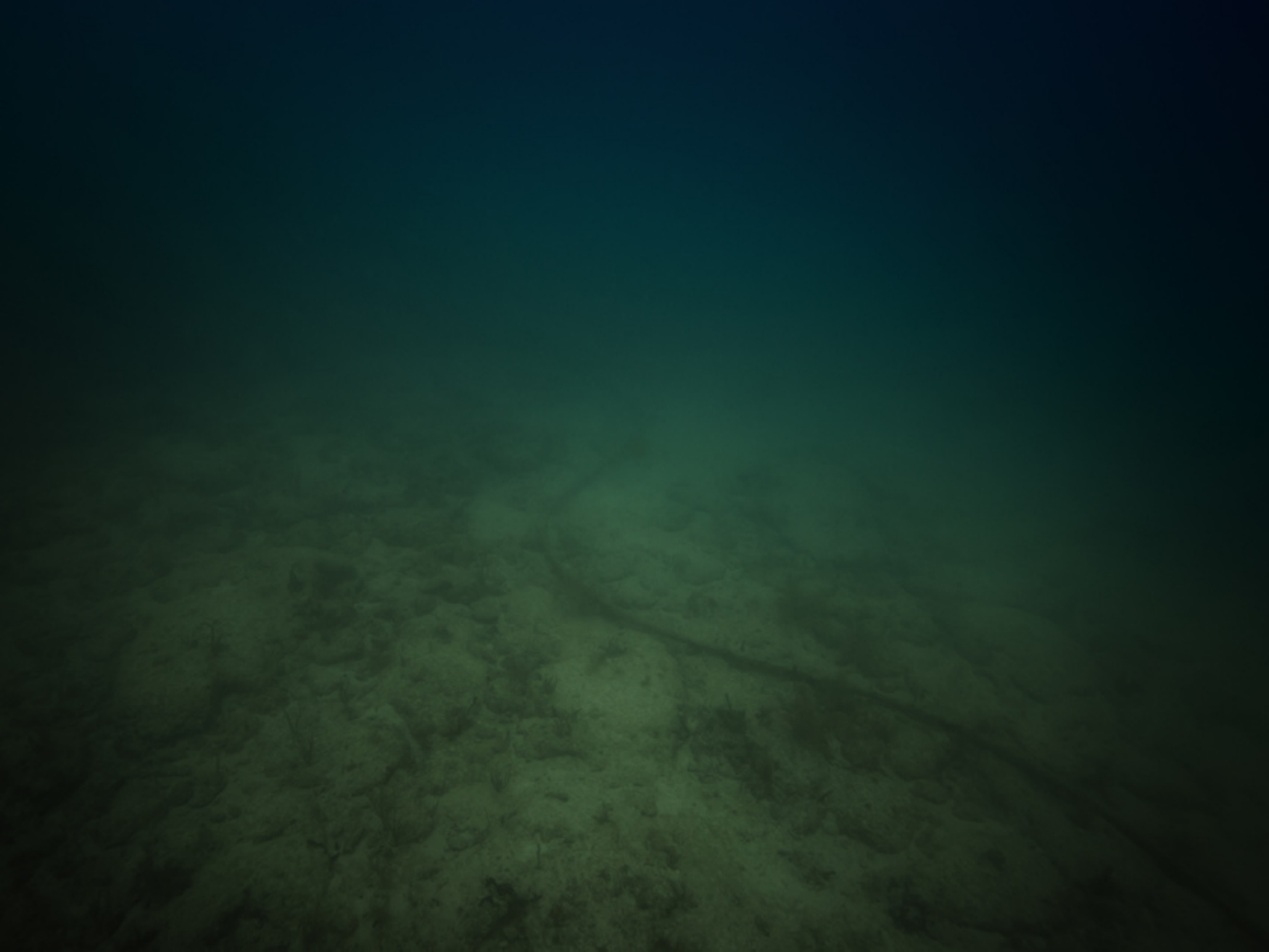} &
\includegraphics[width=4cm,height=3.6cm]{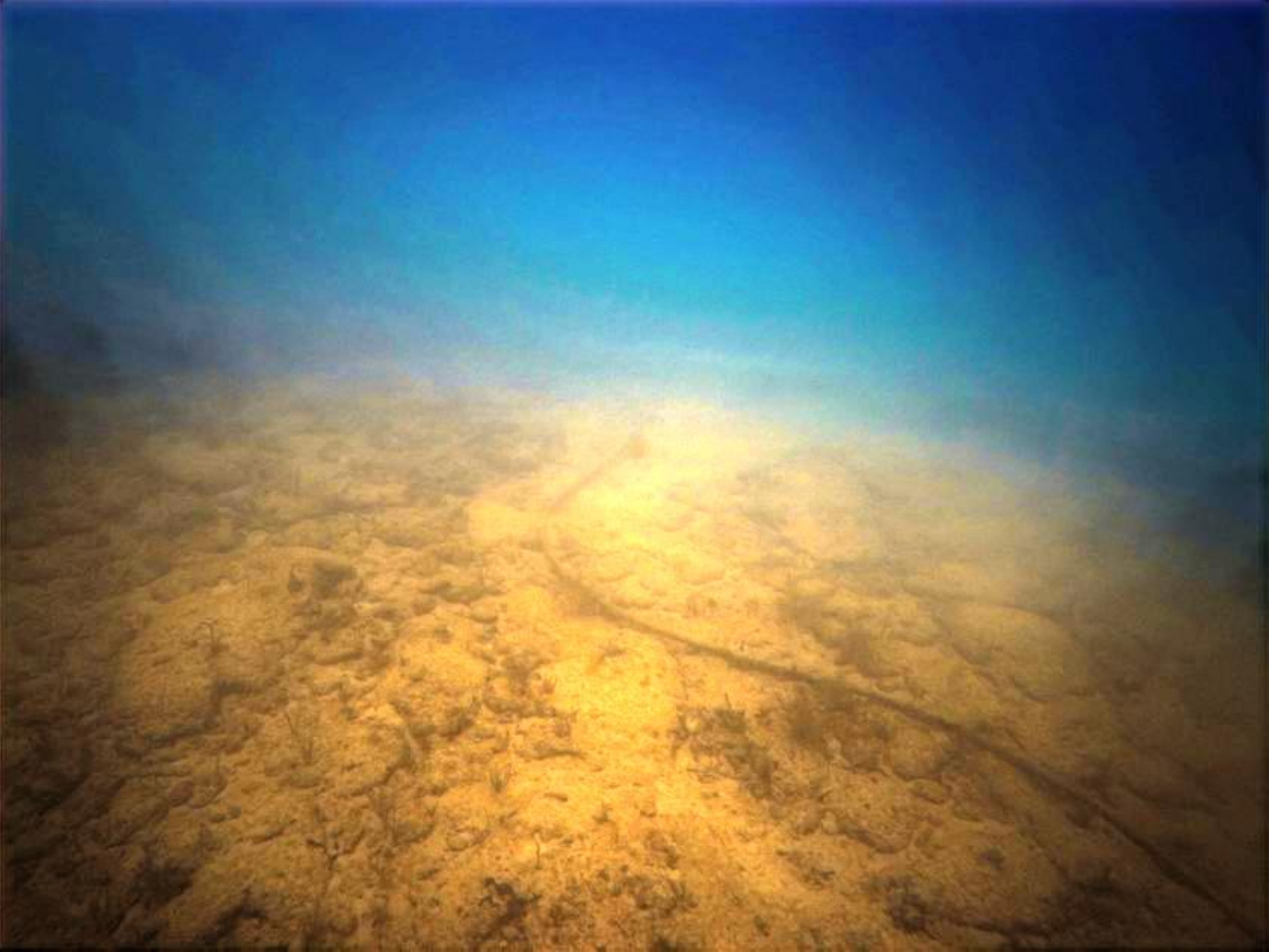}\\
 (a) Input  & (b) Enhanced \\
\end{tabular}
\end{center}

\caption {Enhanced result by the proposed method for a real-world underwater image. As visible, both texture details and color properties are enhanced effectively (best viewed in color on a digital display).
}
\label{fig:sample_front_image}
\end{figure}

Figure~\ref{fig:sample_front_image} shows an example of the enhanced image for a typical real-world underwater image. As visible in Figure~\ref{fig:sample_front_image}(b), our method can accurately recover the underlying color distribution and unveil the suppressed details.

\vspace{2mm}
\noindent {\bf Contributions:} Inspired by deep learning based approaches for low-level visual tasks (\eg, image dehazing by \cite{Cai2016}, image de-raining by \cite{Fu2017}, image depth estimation by \cite{Saeed2017depth}), we design an end-to-end solution for our complex and nonlinear model using a novel convolutional neural network architecture. Our model robustly restores the degraded underwater images and accurately reconstructs underlying colors and appearance. To summarize,
\begin{itemize} 

\item We introduce a novel convolutional neural network model to reconstruct the clear latent underwater image while preserving the original structure and texture by jointly optimizing MSE and SSIM losses. Unlike other mapping function objective minimizing CNN based approaches, our network learns the difference between the degraded underwater image and its clean counterpart.

\item We incorporate a new underwater image synthesis method that is capable of simulating a diverse set of degraded underwater images for data augmentation. To our best knowledge, it is the first underwater image synthesis method that can simulate different underwater types and degradation levels. Our image synthesis can be used as a guide for subsequent network training and full-reference image quality assessments, which calls for the development of new underwater image enhancement methods.

\item Our method is a fully data-driven and end-to-end model. It attains the state-of-the-art performance and generalizes well both on synthetic and real-world underwater images with varying color and visibility characteristics. Hence, our method is suitable for practical applications.
\end{itemize}

\section{Background}

In this section, we provide the physical model of underwater image formation and the reason for using CNN as a prior
\subsection{Underwater Image Formation}

Following the formulation in \cite{Chiang122012}, the underwater image of light after scattering can be expressed as
\begin{equation}
\U_\lambda(x)=\I_\lambda(x)\cdot \T_\lambda(x) + B_\lambda \cdot \big( 1 - \T_\lambda(x) \big)
\label{eq:UW_synthesis}
\end{equation}
where $\U_\lambda(x)$ is the captured underwater image, $\I_\lambda(x)$ is the clear latent image, also called as the scene radiance, that we aim to recover, $B_\lambda$ is the homogeneous global background light, $\lambda$ is the wavelength of the light for the red, green and blue channels, and $x$ is a point in the underwater scene (for clarity, images are denoted in bold capital letters). The medium energy ratio $\T_\lambda(x)$ represents the percentage of the scene radiance reaching at the camera after reflecting from the point $x$ in the underwater scene, which thereby causes color cast and contrast degradation. In other words, $\T_\lambda(x)$ is a function of the wavelength of the light $\lambda$ and the distance $d(x)$ from scene point $x$ to the camera
\begin{equation}
\T_{\lambda}(x) = 10^{-\beta_\lambda d(x)} = \frac{\E_\lambda \big( x,d(x) \big)}{\E_\lambda(x, 0)} = \N_\lambda \big( d(x) \big)
\label{eq:UW_synthesis1}
\end{equation}
where $\beta_\lambda$ is the wavelength-depended medium attenuation coefficient as shown in Figure~\ref{fig:Jerlov}. Assuming the energy of a light beam emanated from $x$ before and after it passes through a transmission medium at a distance of $d(x)$ is $\E_\lambda(x,0)$ and $\E_\lambda \big(x,d(x) \big)$, respectively. The normalized residual energy ratio $\N_\lambda$ corresponds to the ratio of residual energy to initial energy for every unit of distance propagated. Its value varies in water depending on the light wavelength. For example, red light possesses a longer wavelength thus it attenuates faster and gets absorbed more than other wavelengths in water, which results in a bluish tone of most underwater images (see \cite{Chiang122012} for an extended discussion).

\subsection{Formulation as an Image Restoration Task}

To recover the latent image $\I$, the conventional underwater image restoration methods attempt to estimate not only $\I$ but also $B$ and $\T$ from a single underwater image $\U$, which constitutes an underdetermined system as there are fewer equations than the unknowns. Conventional methods also proceed through two main steps. First, they estimate the model parameters of the homogeneous global background light $B$ and the medium energy ratio $\T$. After the model parameter estimation, they reconstruct the latent image $\I$ by inverting the underwater formation model.

Unlike the previous methods, we directly estimate the latent image $\I$ without calculating the homogeneous global background light and the medium energy ratio and we treat the underwater image enhancement as an image restoration task. Instead of direct estimation, our method recovers the residual between the target latent image and the given underwater image in a data-driven and end-to-end manner. Image restoration is an ill-posed problem due to the loss of information, which results in an underdetermined system. To allow a unique solution in the solution space, image restoration methods resort to imposing (often heuristic) regularization schemes or application and domain specific priors.

For a  given the underwater image $\U$, we find the most likely reconstruction of the latent image $\I$ using a maximum a posteriori (MAP) estimator assuming the relation between these two images in terms of a nonlinear function $\U = f(\I)$. This suggests a maximization over the probability distribution of the posterior $p(\I|\U)$ using the Bayes rule
\begin{equation}
p(\I|\U) = p(\U|\I) \frac{p(\I)}{p(\U)}
\label{eq:bayes_prob}
\end{equation}
where $p(\U|\I)$ is the likelihood of observing $\U$ given that $\I$ is the scene radiance and $p(\I)$ is the prior on the latent image. Having a uniform distribution $p(\U)$ on the observations, the maximization of the posterior
\begin{equation}
\begin{split}
\I &= \argmax_\I \left[ p(\I|\U) \right] = \argmax_\I \left[ p(\U|\I) p(\I) \right]
\label{eq:bayesian_prob}
\end{split}
\end{equation}
can be obtained by minimizing the negative log likelihood as
\begin{equation}
\begin{split}
\I & = \argmin_\I \left[-\log  p(\I|\U) \right]\\
&=\argmin_\I \left[-\log \big( p(\U|\I) \big) - \log \big( p(\I) \big) \right] \\
& = \argmin_\I \left[ \big| \U - f(\I) \big|^2 + \alpha \Psi(\I) \right]
\label{eq:minimization_term}
\end{split}
\end{equation}
where the quadratic data fidelity $|\U - f(\I) |^2$ term enforces the observations to be faithful to the degraded version of the estimated latent image, and $\alpha$ is a trade-off parameter between the contributions of the prior and the data fidelity terms.
The regularization prior $\Psi(\I)$ guides the optimization process to the desired output. To model $\Psi(\cdot)$, existing image restoration methods use total variation (\cite{chambolle2004TV}), Gaussian mixture model (\cite{zoran2011GMM}), K-SVD (\cite{elad2006KSVD} and tailored class-specific priors (\cite{anwar2015deblur} $\&$ \cite{anwar2017category}). Reformulating the underwater image enhancement as an image restoration problem with an explicit regularization prior offers several benefits; i) The exact parameterization of the prior $\Psi(\cdot)$ can be unknown in solving Eq~\eqref{eq:minimization_term}. ii) Alternative image priors can be exploited jointly. iii) Such priors can be plugged into other similar inverse problems.

\subsection{Why CNN?}

Our choice for image restoration and regularization in Eq~\eqref{eq:minimization_term} is a cascade of hierarchical nonlinear filtering with progressively increasing local receptive fields, which are modeled by the convolutional neural networks without spatial pooling. Our model imposes a natural bottleneck to the residual between the latent and underwater images as its number of parameters is much smaller than the number of input pixels yet its patch filters allow guiding its enhancement factors based on the blur and color within local patches. Furthermore,
\begin{itemize}
\item CNN provides a strong modeling capability of the distortions that we aim to compensate between the latent and original underwater images and facilitates discriminative prior learning.
\item It models the external priors by learning the relationship between the underwater image and the ground truth, while many existing methods attempt to model internal priors relying on the underwater image statistics and features \eg, ODM \cite{Li32016} and are complementary to our method. By combining our approach with ODM \cite{Li32016} is expected to improve the performance.
\item Inference on a CNN can be made efficiently by exploiting the parallel processing platforms.
\end{itemize}

\begin{figure}[!t]
\begin{center}
\includegraphics[width=\linewidth]{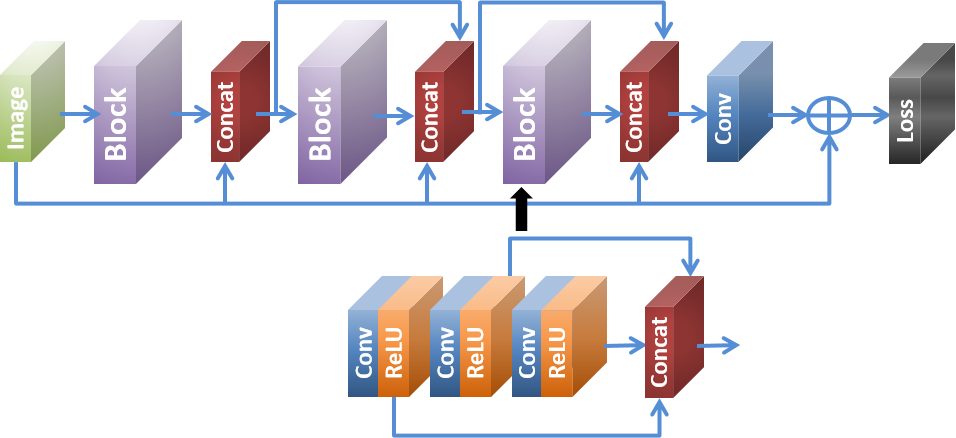}
\end{center}
\caption{Our UWCNN model where ``Conv" are the convolutional layers, ``Concat" are the stacked convolutional layers, ``ReLU" is the rectified linear unit.
}
\label{fig:Network_structure}
\end{figure}

\section{Our Proposed Model}
Here, we discuss the details of the proposed underwater image enhancement fully Convolutional Neural Network (UWCNN) and then present a post-processing stage to further improve our enhanced result.

\subsection{Network Architecture}
Figure~\ref{fig:Network_structure} shows the architecture of our underwater image enhancement model UWCNN. It is a densely connected fully convolutional network that is inspired by the recent densely connected network models in object classification \cite{huang2016densely}. In the following section, we present its basic building blocks and hyperparameters. The input to our network is a RGB underwater image $\U$.

\textbf{Residuals:} Unlike the conventional approaches that directly predict the clean latent image $\I$ by learning the mapping function $\I = f^{-1}(\U)$, we allow our network to learn the difference between the synthetic underwater image and its clean counterpart. Note that such a synthetic image generation task is a nontrivial objective for underwater image enhancement and restoration field, and it will be discussed in detail in Section~\ref{sec:data_generation}. As underwater image and its feature maps in the subsequent layers are processed through many convolutional filters before reaching the final loss layer. Although our network is not intentionally very deep, there is still a possibility of vanishing or exploding gradients as mentioned in \cite{bengio1994vanishing}. To avoid such issues during the training iterations, we enforce learning the residue by adding the input of the network \ie $\U$ to the output of the network \ie $\Delta(\U, \theta)$ (see below) before loss function as
\begin{equation}
\I = \U + \Delta ( \U, \theta )
\label{eq:latent_objective}
\end{equation}
where ``+'' is the element-wise addition operation.

\textbf{Blocks:}
The UWCNN has a modular architecture composed of blocks having same structure and components. Suppose $r$ and $c$ are the notation for ReLU and convolution, then the first operation of convolution and ReLU pair, in the $l$-th block, is given by
\begin{equation}
z_{l, 0} = r \big( c(\U ); \theta_{l, 0} \big)
\end{equation}
where $z_{l, 0}$ is the output of the first convolution-ReLU pairs of $l$-th residual block and $\theta_{l, 0}$ is a set of weights and biases associated with it. By composing the series of convolution-ReLU pairs, we obtain
\begin{equation}
z_{l, n} = r \big( c( \ldots r( c( \U; \theta_{l, 0}) )~\ldots ); \theta_{l, n} \big).
\end{equation}
The output of the $l$-th block is obtained by concatenating along third dimension of each individual convolution-ReLU pairs output $z$'s and input image $\U$ as
\begin{equation}
b_l = h( z_{l, 0};~\ldots;~z_{l, n};~\U).
\end{equation}
The output of the $l$-th block is obtained by concatenating along the third dimension of each convolution-ReLU pairs output $z$'s and input image $\U$ as
\begin{equation}
b_{l+1} = h( z_{l+1, 0};~\ldots;~z_{l+1, n};~\U;~b_l).
\end{equation}
Finally, we chain all blocks and the output of this chain is convolved with a final convolution layer with parameters $\theta_{l+m,n}$ to  predict the component as $ \Delta ( \U, \theta ) = c( b_{l+m},\theta_{l+m,n})$.

\textbf{Network Layers:} Our fully convolutional network consists of three different layers indicated by different colors as shown in Figure~\ref{fig:Network_structure}. The first type is of convolutional layers represented by ``Conv'' in Figure~\ref{fig:Network_structure}, which consists of 16 convolutional kernels of size $3\times3\times3$ to produce 16 output feature maps for the first layer, while subsequent convolutional layers produce 16 maps each using $3\times3\times16$ filters.

The second type of the layers is activation layers, also known as ``ReLU'', to introduce the nonlinearity. The third type is ``Concat'' layers, which is used to concatenate all the convolutional layers after each block. The last convolutional layer estimates the final output of the network, which is the latent image.

\textbf{Dense Concatenation:} We stack all convolutional layers at the end of each block. This technique is different from DenseNet provided in \cite{huang2016densely}, where each convolutional layer is connected with other convolutional layers in the same block. Furthermore, we do not use any fully connected layers or batch normalization steps, which makes our network memory efficient and fast. In addition, we feed the input image to every block as can be seen from connections of Figure~\ref{fig:Network_structure}. The stacking of the convolutional layers with input data reduces the need for a very deep network, and this property of stacking layers can be attributed to the superior performance of our proposed system. In summary, our network is unique since:
\begin{itemize}

 \item the input image is applied to all blocks, and
\item it contains only the fully-convolutional layers without any batch-normalization steps.
\end{itemize}

\textbf{Network Depth:} Our network is of modular structure and consists of three blocks where each block is again composed of three convolutional layers. We have a single convolutional layer at the end of the network, hence, making the full depth of our network only ten layers. This makes our model computationally inexpensive and highly practical in training as well as during prediction.

\textbf{Reducing Boundary Artifacts:} In end-to-end low-level vision tasks, the output of the system is needed to be equal to the input of the system. This requirement sometimes results in boundary artifacts. To avoid this phenomenon, we enforce two strategies:
\begin{itemize}

\item we do not use any pooling layers in our network, and
\item we add zeros before each convolutional layer.
\end{itemize}
As a consequence, the final output image of UWCNN network is almost artifacts-free around the boundaries and is of the same size as the input image.

\subsection{Network Loss}
\label{sec:network_loss}

Since we aim to reconstruct an image, we compute the objective function loss pixel-wise. We use the $\ell_2$ loss, \ie, MSE (Mean Square Error) as in our observations it can well preserve the sharpness of edges and details. Blurring edges results in large errors. We add the estimated residual to the input underwater image, then compute the MSE loss as
\begin{equation}
\label{equ_8}
  L_{MSE} = \frac{1}{M} \sum_{i=1}^M \left| \bigg[ \U(x_i) + \Delta \Big( \U(x_i), \theta(x_i) \Big) \bigg] - \I^*(x_i) \right|^2
\end{equation}
where $\U(x_i) + \Delta( \U(x_i), \theta(x_i) ) = \I(x_i)$ is the estimated latent image pixel value at $x_i$, $i=1,..,M$ as described in Eq.~\eqref{eq:latent_objective} and $\I^*_i$ is the corresponding ground truth image in the training dataset.

In addition, we include the SSIM loss \cite{Zhao2017} in our objective function to impose the structure and texture similarity on the latent image. We use the gray images to compute SSIM scores. For each pixel $x$, the SSIM value is computed within a $13\times13$ image patch around the pixel as
\begin{equation}
\label{equ_9}
  S\!S\!I\!M(x) = \frac{2\mu_{I*}(x) \mu_I(x) + c_1}{ \mu_{I*}^2(x)+\mu_{I}^2(x) + c_1 }\cdot\frac{2\sigma_{I*I}(x) + c_2}{\sigma_{I*}^2(x) + \sigma_{I}^2(x) + c_2}
\end{equation}
where $\mu_{I}(x)$ and $\sigma_{I}(x)$ correspond to the mean and standard deviation of the image patch from the latent image $\I$, similarly, $\mu_{I*}(x)$ and $\sigma_{I*}(x)$ are for the patch from the ground truth image $\I^*$. The cross-covariance $\sigma_{I*I}(x)$ is computed between the patches from $\I$ and $\I^*$ for the pixel $x$. We set constants $c_{1}=0.02$ and $c_{2}=0.03$ based on the default in SSIM loss. Our model is insensitive to these defaults. Still, we fix them for a fair comparison.  Finally, the SSIM loss is expressed as
\begin{equation}
\label{equ_10}
L_{SSIM} = 1 - \frac{1}{M} \sum_{i=1}^M S\!S\!I\!M (x_i).
\end{equation}
The final loss function $L$ is the aggregation of MSE and SSIM losses
\begin{equation}
\label{equ_11}
\begin{aligned}
  L = L_{MSE} + L_{SSIM}.
\end{aligned}
\end{equation}
At last, we learn the residual mappings between the degraded input and the latent result. Instead of the classical stochastic gradient descent (SGD) optimizer (see \cite{LeCun1998}), we use ADAM (see \cite{Kingma2014}) to update our UWCNN weights iteratively. ADAM combines the advantages of AdaGrad (see \cite{Duchi2011}) that works well with sparse gradients, and RMSProp (see \cite{Tieleman2012}) that works well in on-line and non-stationary settings, which is relatively easy to configure where the default configuration parameters do well on most problems. Following the popular settings in low-level vision task networks, we assign the learning rate to 0.0002, ADAM's internal parameters $\beta_1$ to 0.9 and $\beta_2$ to 0.999 (see \cite{Kingma2014} for their definition), and fix the learning rate in the entire training procedure.

\subsection{Post-processing}

\begin{figure*}[!t]
\begin{center}
\begin{tabular}{ccc}

\includegraphics[width=4.5cm,height=10cm]{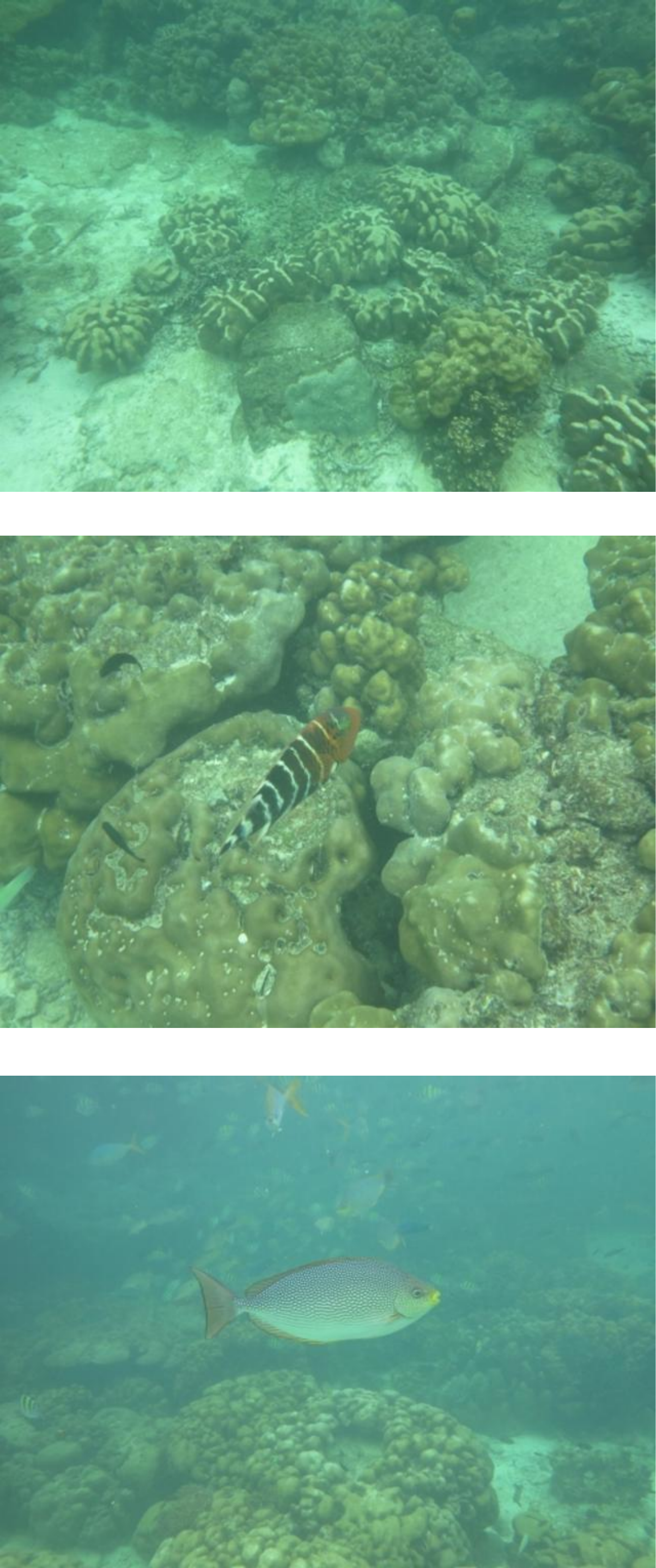}&
\includegraphics[width=4.5cm,height=10cm]{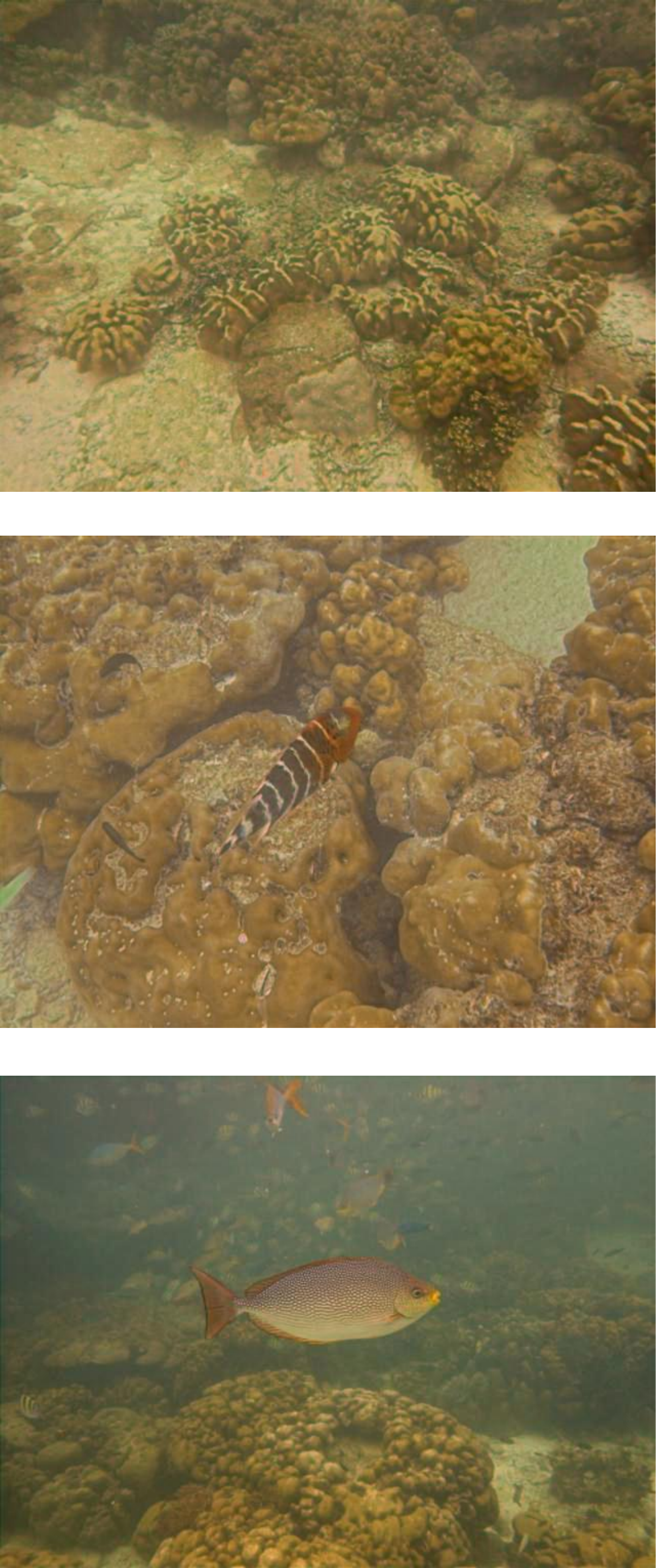}&
\includegraphics[width=4.5cm,height=10cm]{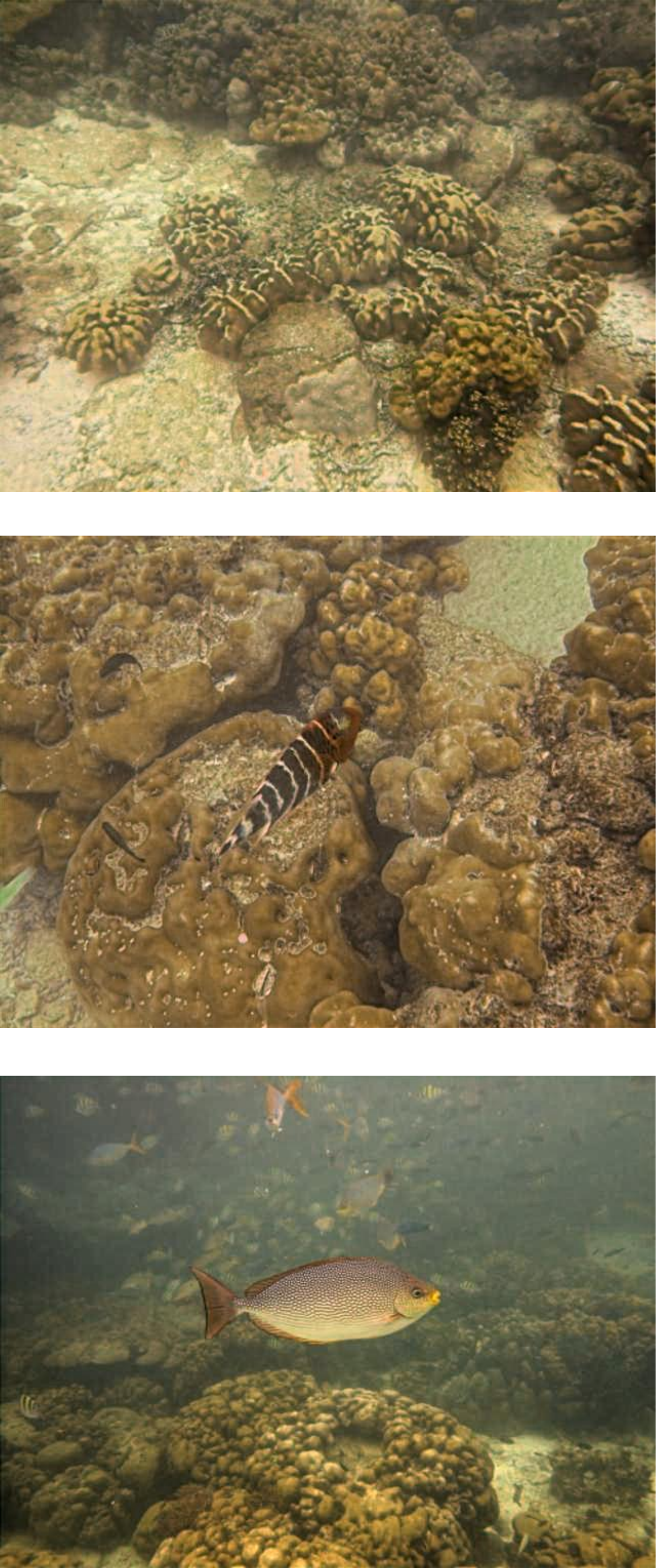}\\
(a)&(b)&(c)\\
\end{tabular}
\end{center}
\caption{Sample results for qualitative assessment. (a) Original real-world underwater images, (b) results of UWCNN, (c) results of UWCNN+. As visible, our methods remove the greenish tone while reconstructing accurate and vivid latent images.}
\label{fig:real_images_show}
\end{figure*}

UWCNN generates enhanced images without color casts and exceptional visibility. However, due to the limitation of our training data pairs (an indoors image as the latent image and a synthesized image from the indoors image using the aforementioned underwater image formation model as the corresponding underwater image), the enhanced images have lower dynamic range. In practice, one would expect the enhanced results to have vivid colors and higher contrast.

To solve this issue, we employ a simple yet effective adjustment as a post-processing stage. We denote UWCNN with post-processing as UWCNN+. The image is first transformed to HSI color space. Then, the ranges of its saturation and intensity components in the HSI color space are normalized to $[0,1]$ as
\begin{equation}
\label{equ_11}
y_{out} = \frac{y_{in} - y_{min} }{ y_{max} - y_{min} }
\end{equation}
where $y_{max}$ and $y_{min}$ are the maximum and minimum saturation or intensity values in the UWCNN image. To avoid the effect of a single pixel having too high or too low values on the selection of maximum and minimum saturation or intensity values, we calculate the histogram distribution of saturation or intensity and select $0.2\%$ (frequency) as determinant. This means ignoring all the pixels values that frequency $<0.2\%$ and only applying the normalization method to the  frequency $\geq 0.2\%$ pixel values.  $0.2\%$ is a heuristic value based on extensive experiments.  After this simple saturation and intensity normalization, we transform the modified result back to RGB color space.

Sample results are shown in Figure~\ref{fig:real_images_show}. As visible, UWCNN removes effectively the dominant greenish color distortion in these real-world underwater images and improves significantly the contrast while preserving natural look and authenticity of the images. Compared to UWCNN, the saturation and intensity normalization in UWCNN+ further improves the contrast and brightness, unveiling more details. Our methods do not introduce extra colors as some other existing methods either.

\section{Experimental Evaluations}
\label{sec:experiments}

To evaluate our method, we perform qualitative and quantitative comparisons with the recent state-of-the-art underwater image enhancement methods on both synthetic and real-world underwater images. These methods include  UDCP by \cite{Drews152016}, RED by \cite{Galdran142015}, ODM by \cite{Li32016} and UIBLA by \cite{Peng2017}. We run the source codes provided by the corresponding authors with the recommended parameter settings to produce the best results for an objective evaluation. Since WaterGAN \cite{Li2017} is only applicable to specific scenarios, its results are not competitive and therefore not reported in this paper.  For real-world images where the light-attenuation coefficients are not available, we apply each of the ten UWCNN models we learned and present the results that are visually more appealing. This process can be improved by using a classification stage to choose the best model, which we plan to explore as future work. For synthetic data, we present the results without post-processing since the models are derived from the synthetic data thus no intensity and saturation normalization is required. At last, we conduct an ablation study to demonstrate the effect of each component of our network.

Unlike the high-level visual analytics tasks where large training datasets are often available, the underwater image enhancement depends on the synthetic data. Next, we explain how we generate the synthetic underwater images. To the best of our knowledge, this is the first physical model based underwater image synthesis method that can simulate a diverse set of water types and degradation levels, which is a significant contribution for the development of data-driven underwater image enhancement techniques.

\subsection{Underwater Image Synthesis}
\label{sec:data_generation}

\begin{table*}[!t]
\centering
\caption{$N_\lambda$ values for synthesizing ten underwater image types.}
\begin{tabular}{|l|c|c|c|c|c|c|c|c|c|c|}
\hline
Types  & I      & IA     & IB    & II     & III & 1  & 3   & 5 & 7 & 9 \\ \hline \hline
blue   &0.982   & 0.975  & 0.968 & 0.94   & 0.89    & 0.875  & 0.8    & 0.67    & 0.5   &   0.29\\
green  &0.961   & 0.955  & 0.95  & 0.925  & 0.885   & 0.885  & 0.82   & 0.73    & 0.61  &   0.46\\
red    &0.805   & 0.804  & 0.83  & 0.8    & 0.75    & 0.75   & 0.71   & 0.67    & 0.62  &   0.55\\
\hline
\end{tabular}
\label{table:Nrer_values}
\end{table*}

\begin{figure*}[!htbp]
\begin{center}
\begin{tabular}{c@{ } c@{ } c@{ } c@{ } c@{ } c}
\includegraphics[width=.16\textwidth]{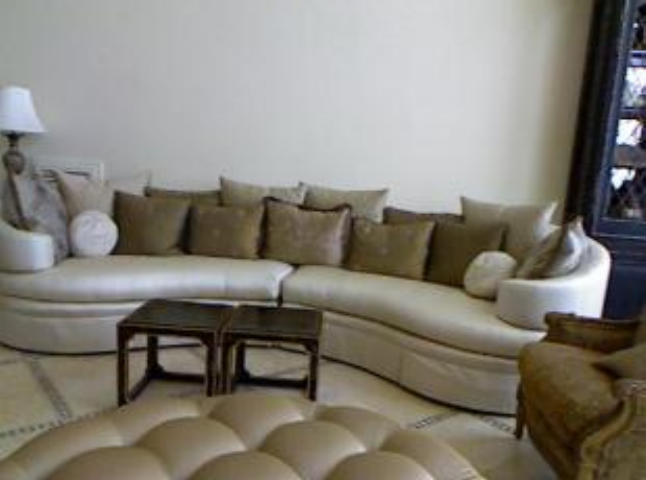}&
\includegraphics[width=.16\textwidth]{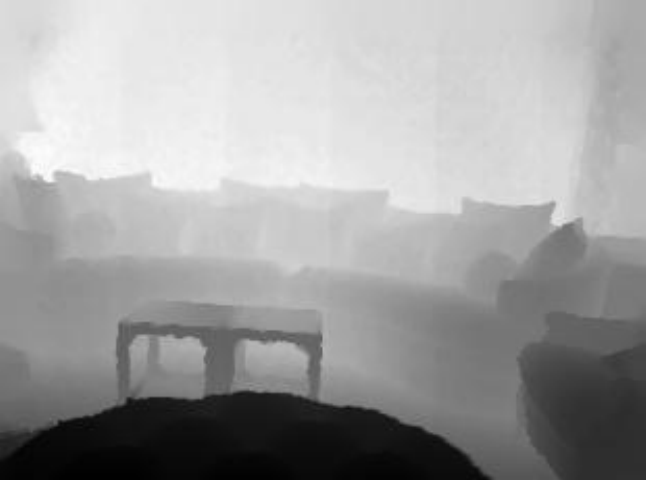}&
\includegraphics[width=.16\textwidth]{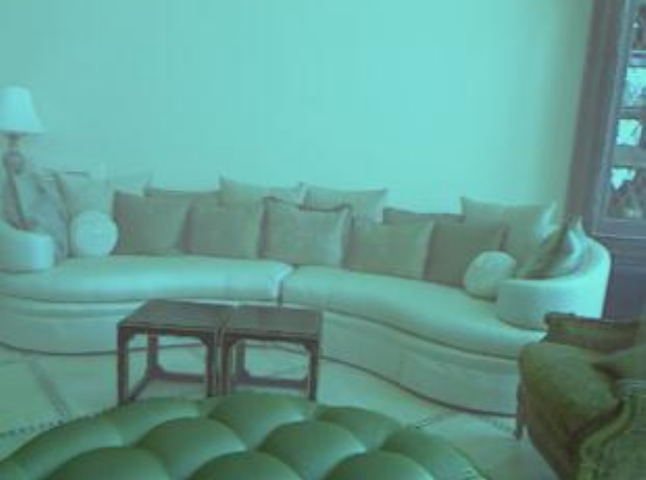}&
\includegraphics[width=.16\textwidth]{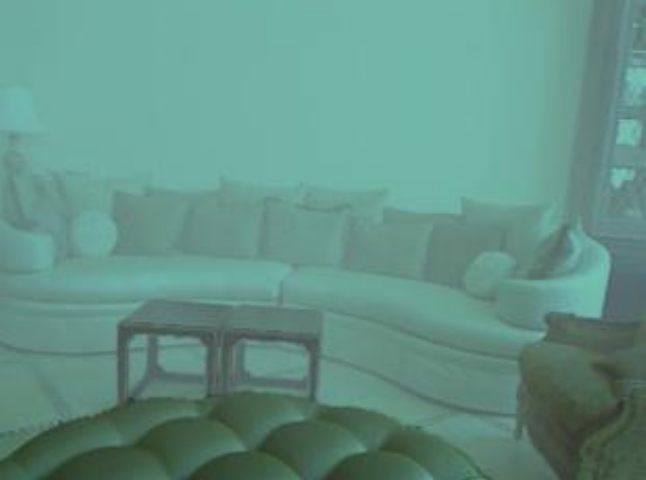}&
\includegraphics[width=.16\textwidth]{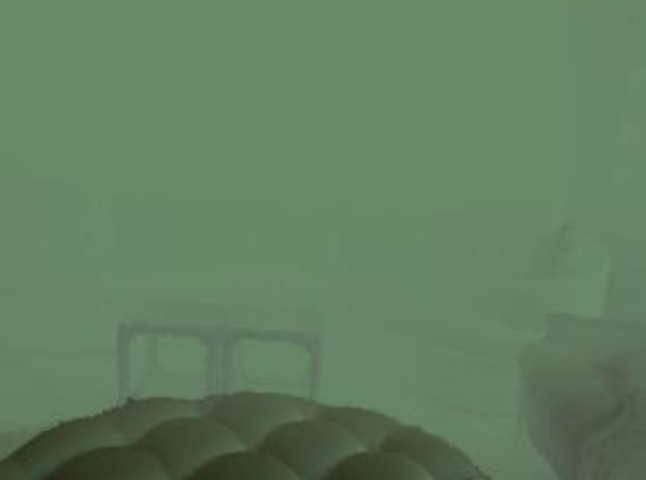}&
\includegraphics[width=.16\textwidth]{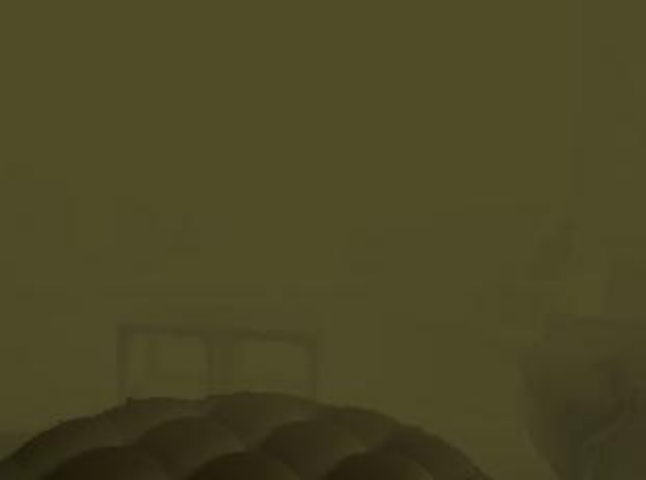}\\
 clear image & depth map & Type-1 & Type-3 & Type-5 & Type-7\\
\includegraphics[width=.16\textwidth]{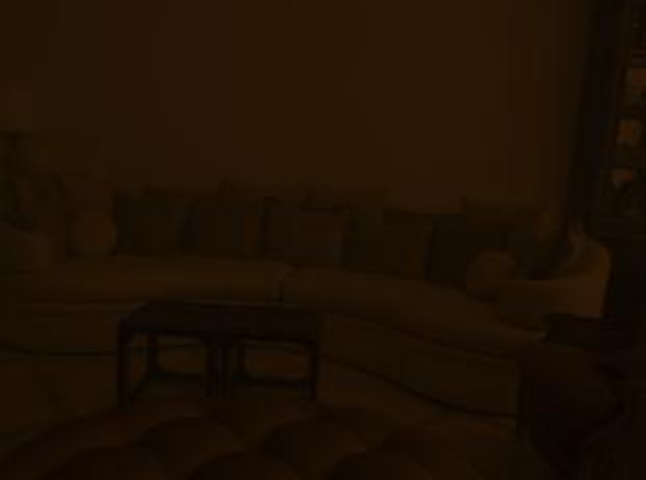}&
\includegraphics[width=.16\textwidth]{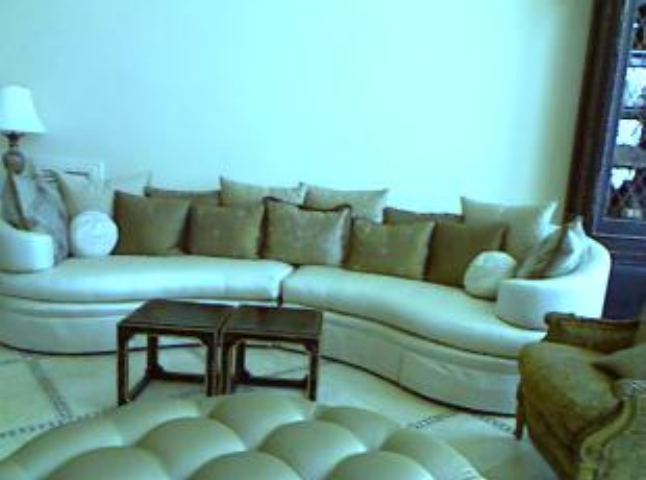}&
\includegraphics[width=.16\textwidth]{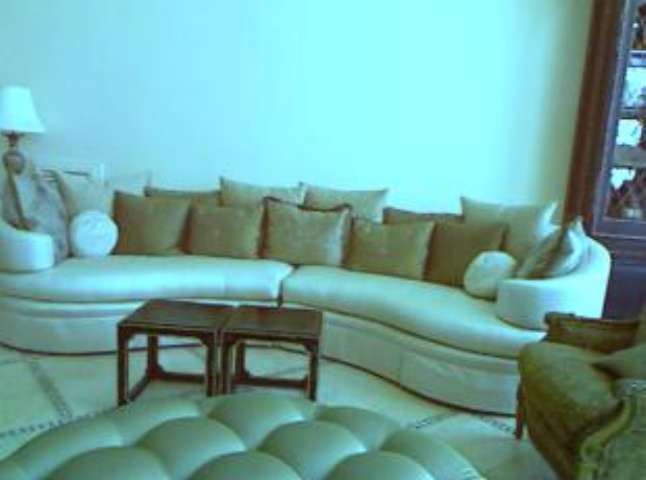}&
\includegraphics[width=.16\textwidth]{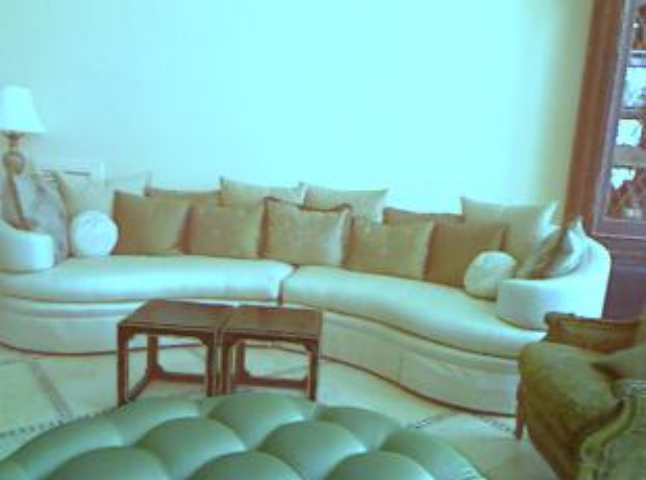}&
\includegraphics[width=.16\textwidth]{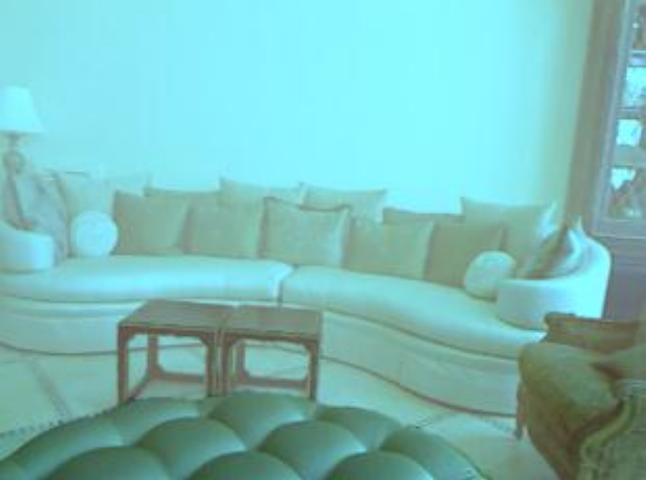}&
\includegraphics[width=.16\textwidth]{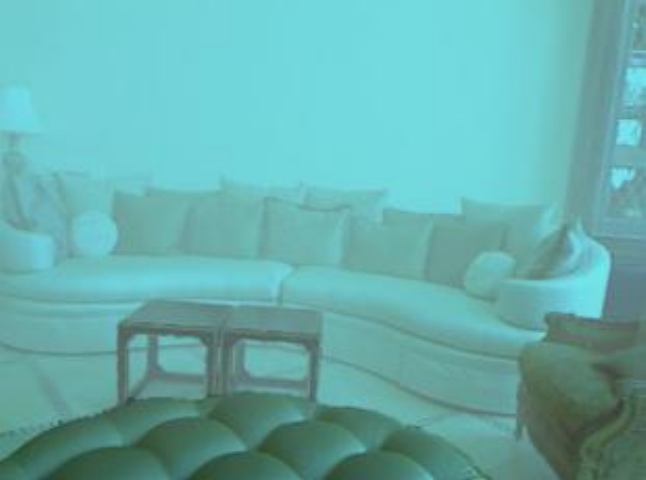}\\
Type-9 & Type-I & Type-IA & Type-IB & Type-II & Type-III\\
\end{tabular}
\end{center}
\caption{Ten types of synthesized underwater images from the NYU-v2 RGB-D dataset~\cite{Silberman2012} using a sample image and its depth map.}
\label{fig:sample_image}
\end{figure*}

To synthesize the training set for our UWCNN model, we use the attenuation coefficients described in \cite{Jaffe1976} for the different water types of  oceanic and  coastal classes (\ie, I, IA, IB, II, and III for open ocean waters, and 1, 3, 5, 7, and 9 for coastal waters).  As mentioned before, Type-I is the clearest and Type-III is the most turbid open ocean water. Similarly, for coastal waters, Type-1 is the clearest and Type-9 is the most turbid. We apply Eq~\eqref{eq:UW_synthesis} and Eq~\eqref{eq:UW_synthesis1} to build ten types of underwater image datasets using the RGB-D NYU-v2 indoor dataset of \cite{Silberman2012}, which consists of 1449 images. We select the first $1000$ clean images for the training set and the remaining $449$ clean images for the validation (test) set.

To synthesize an underwater image, we generate a random homogeneous global atmospheric light $0.8< B_\lambda < 1$. Then, we vary the depth $d(x)$ from 0.5m to 15m, which is followed by the selection of the corresponding $N_\lambda$ values of the red, green, and blue channels for the water types presented in Table~\ref{table:Nrer_values}. For each image, we generate five underwater images based on random $B_\lambda$ and depth $d(x)$; therefore, we obtain a training set of 5000 and a validation set of 2495 samples. For computational efficiency, we resize these images to $310\times230$. In total, we synthesize ten underwater image datasets and train ten UWCNN models for each type.

Figure~\ref{fig:sample_image} shows these ten different types of underwater images for a sample. It is evident that the underwater images of Type-I, Type-IA and Type-IB are similar in their physical appearance and characteristics. Thus, we select a total of eight models out of ten to display the results on synthetic underwater images.

 \begin{figure*}[!t]
  \centering
\begin{minipage}[b]{0.135\linewidth}
  \centering
  \centerline{\includegraphics[width=2.3cm,height=16cm]{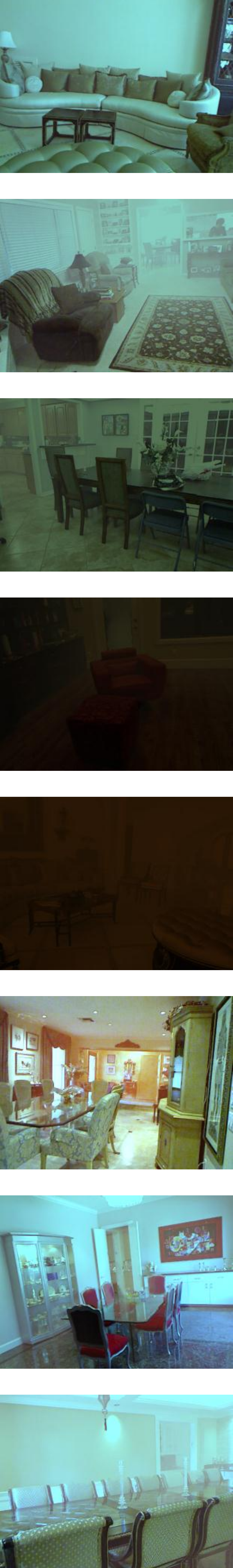}}
  \centerline{(a) Raws}\medskip
\end{minipage}
\begin{minipage}[b]{0.135\linewidth}
  \centering
  \centerline{\includegraphics[width=2.3cm,height=16cm]{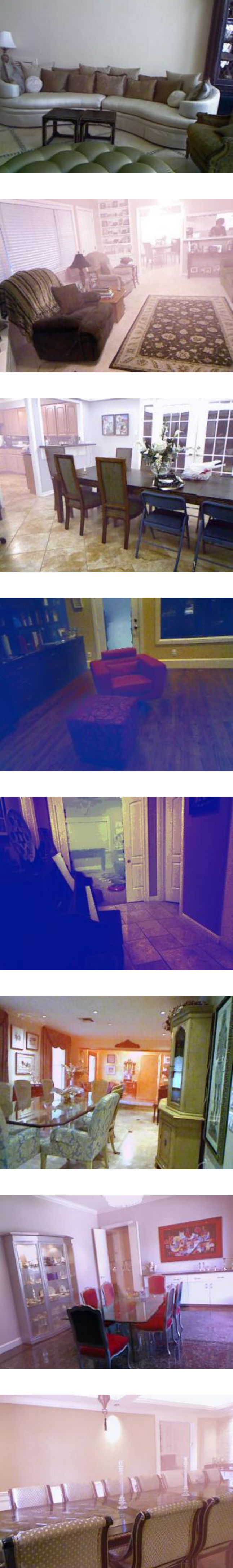}}
  \centerline{(b) RED }\medskip
\end{minipage}
\begin{minipage}[b]{0.135\linewidth}
  \centering
  \centerline{\includegraphics[width=2.3cm,height=16cm]{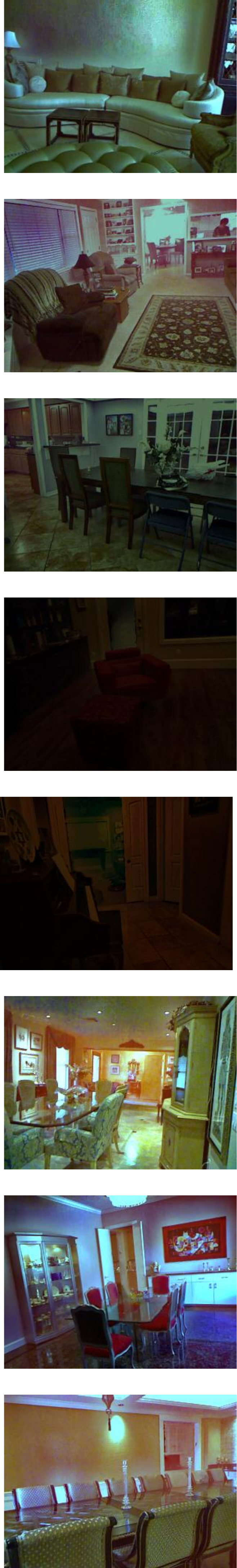}}
  \centerline{(c) UDCP}\medskip
\end{minipage}
\begin{minipage}[b]{0.135\linewidth}
  \centering
  \centerline{\includegraphics[width=2.3cm,height=16cm]{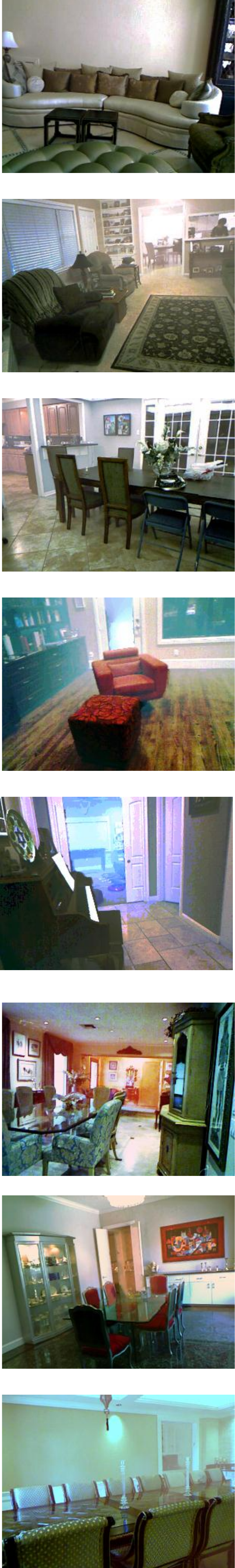}}
  \centerline{(d) ODM}\medskip
\end{minipage}
\begin{minipage}[b]{0.135\linewidth}
  \centering
  \centerline{\includegraphics[width=2.3cm,height=16cm]{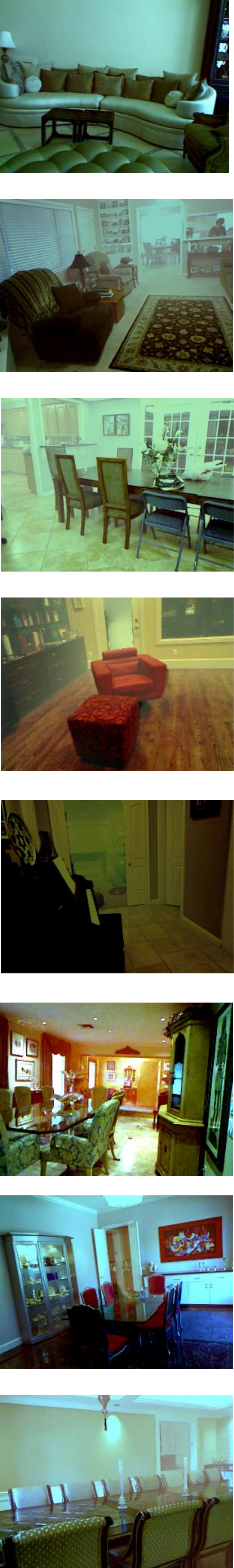}}
  \centerline{(e) UIBLA}\medskip
\end{minipage}
\begin{minipage}[b]{0.135\linewidth}
  \centering
  \centerline{\includegraphics[width=2.3cm,height=16cm]{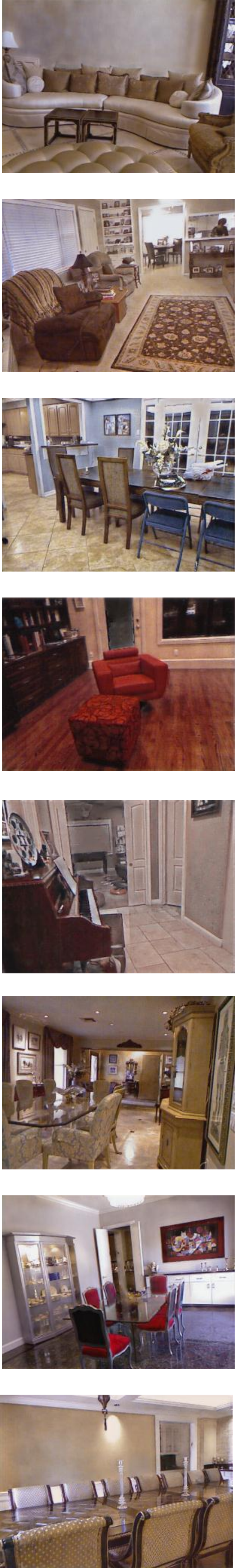}}
  \centerline{(f) UWCNN }\medskip
\end{minipage}
\begin{minipage}[b]{0.135\linewidth}
  \centering
  \centerline{\includegraphics[width=2.3cm,height=16cm]{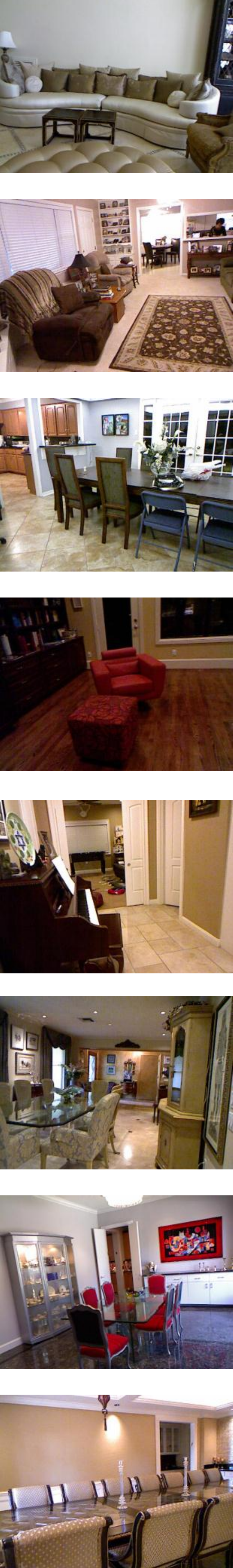}}
  \centerline{(g) GT }\medskip
\end{minipage}
\caption{Qualitative comparisons for samples from the test dataset. Our network removes the light absorption effects and recovers the original colors without any artifacts. The types of underwater images in the first column from top to bottom are Type-1, Type-3, Type-5, Type-7, Type-9, Type-I, Type-II, and Type-III.}
\label{fig:synthetic_results}
\end{figure*}

\begin{table*}[!t]
\begin{center}
\caption{Quantitative evaluations on the test set. As seen, our method achieves the best scores in all metrics on all underwater image types. For the MSE, a lower score is better. For PSNR and SSIM, higher scores are better. }
\begin{tabular}{ccccccccc} \hline
 & \textbf{Types}   & \textbf{RAW} & \textbf{RED} & \textbf{UDCP} & \textbf{ODM} & \textbf{UIBLA} & \textbf{UWCNN} \\ \hline
       &1   & 2367.3  & 3489.7  & 2062.3  & 2508.6  & 2812.6 &  \textbf{587.70}  \\
	   &3   & 2676.5  & 4953.2  & 3380.6  & 3130.1  & 3490.1 &  \textbf{747.50}  \\
       &5   & 4851.2  & 8385.8  & 6708.9  & 3488.9  & 4563.7 &  \textbf{1295.1} \\
  MSE  &7   & 7381.1  & 9809.8  & 8591.6  & 5337.1  & 6737.9 &  \textbf{2974.1} \\
       &9   & 9060.6  & 5952.3  & 9500.1  & 10634.0 & 8433.1 &  \textbf{4121.5} \\
	   &I   & 1449.0  & 936.9   & 1020.7  & 1272.0  & 1492.2 &  \textbf{209.70}  \\
       &II  & 941.9   & 851.3   & 1466.0  & 1401.9  & 1141.4 &  \textbf{251.60}  \\
       &III & 1851.0  & 2240.0  & 2337.6  & 1701.1  & 1697.8 &  \textbf{456.40}  \\ \hline

	   &1    & 15.535  & 15.596  & 15.757  & 16.085  & 15.079 &   \textbf{21.790}  \\
	   &3    & 14.688  & 12.789  & 14.474  & 14.282  & 13.442 &   \textbf{20.251}  \\
	   &5    & 12.142  & 11.123  & 10.862  & 14.123  & 12.611 &   \textbf{17.517}  \\
 PSNR  &7    & 10.171  & 9.991   & 9.467   & 12.266  & 10.753 &   \textbf{14.219}  \\
	   &9    & 9.502   & 11.620  & 9.317   & 9.302   & 10.090 &   \textbf{13.232}  \\
	   &I    & 17.356  & 19.545  & 18.816  & 18.095  & 17.488 &   \textbf{25.927}  \\
	   &II   & 20.595  & 20.791  & 17.204  & 17.610  & 18.064 &   \textbf{24.817}  \\
	   &III  & 16.556  & 16.690  & 14.924  & 16.710  & 17.100 &   \textbf{22.633}  \\ \hline
	
       &1     & 0.7065  & 0.7406  & 0.7629  & 0.7240  & 0.6957  & \textbf{0.8558}\\
	   &3     & 0.5788  & 0.6639  & 0.6614  & 0.6765  & 0.5765  & \textbf{0.7951}\\
       &5     & 0.4219  & 0.5934  & 0.4269  & 0.6441  & 0.4748  & \textbf{0.7266}\\
 SSIM  &7     & 0.2797  & 0.5089  & 0.2628  & 0.5632  & 0.3052  & \textbf{0.6070}\\
	   &9     & 0.1794  & 0.3192  & 0.1624  & 0.4178  & 0.2202  & \textbf{0.4920}\\
       &I     & 0.8621  & 0.8816  & 0.8264  & 0.8172  & 0.7449  & \textbf{0.9376}\\
	   &II    & 0.8716  & 0.8837  & 0.8387  & 0.8251  & 0.8017  & \textbf{0.9236}\\
	   &III   & 0.7526  & 0.7911  & 0.7587  & 0.7546  & 0.7655  & \textbf{0.8795}\\ \hline
\end{tabular}
\end{center}
\end{table*}

\subsection{Network Implementation and Training}
\label{sec:trainging}

For training, the input to our network is the synthetic images of size $310\times230$ generated from the NYU-v2 RGB-D dataset without any augmentation or preprocessing. We trained our model using ADAM (see \cite{Kingma2014}) and set the learning rate to 0.0002, $\beta_{1}$ to 0.9, $\beta_{2}$ to 0.999. The batch size is set to 16. It takes around three hours to optimize a model over 20 epochs. We use TensorFlow as the deep learning framework on an Inter(R) i7-6700k CPU, 32GB RAM, and a Nvidia GTX 1080 Ti GPU.

\begin{figure*}[!t]
\centering
\begin{tabular}{ccccccc}
\includegraphics[width=.12\textwidth]{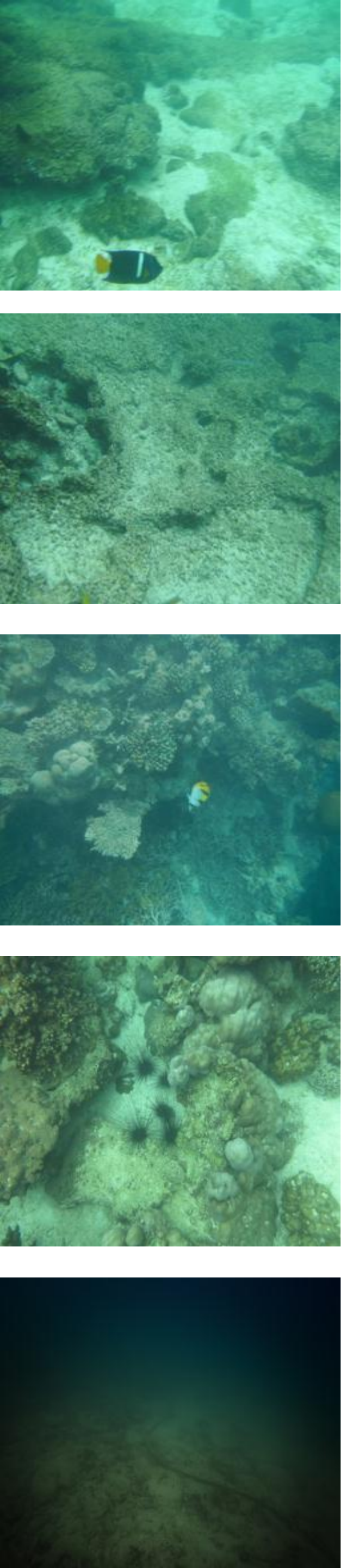}&
\includegraphics[width=.1205\textwidth]{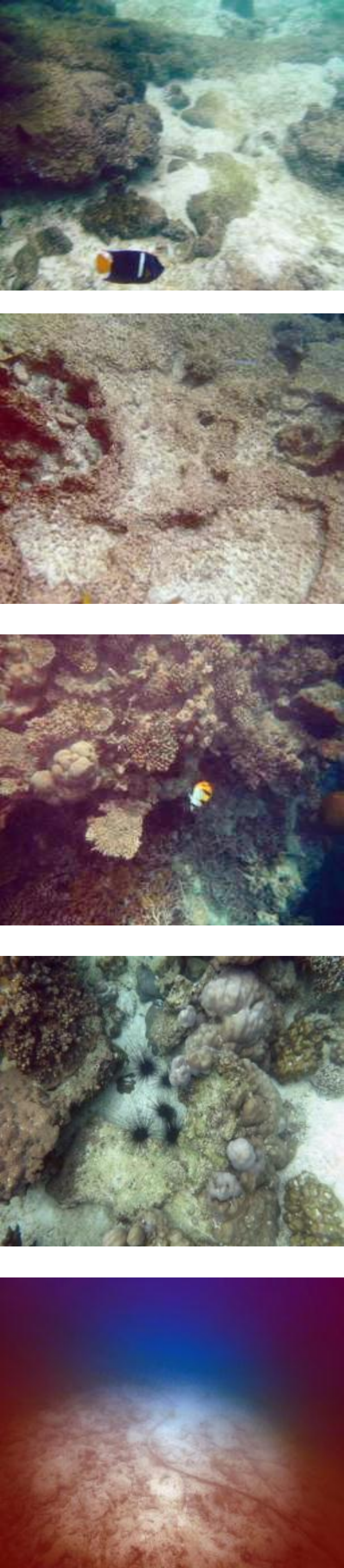}&
\includegraphics[width=.123\textwidth]{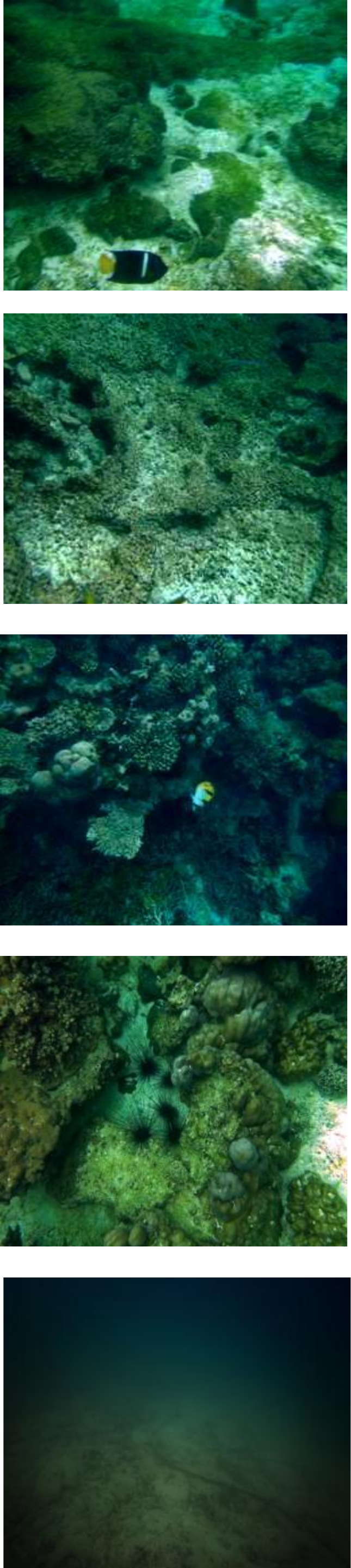}&
\includegraphics[width=.122\textwidth]{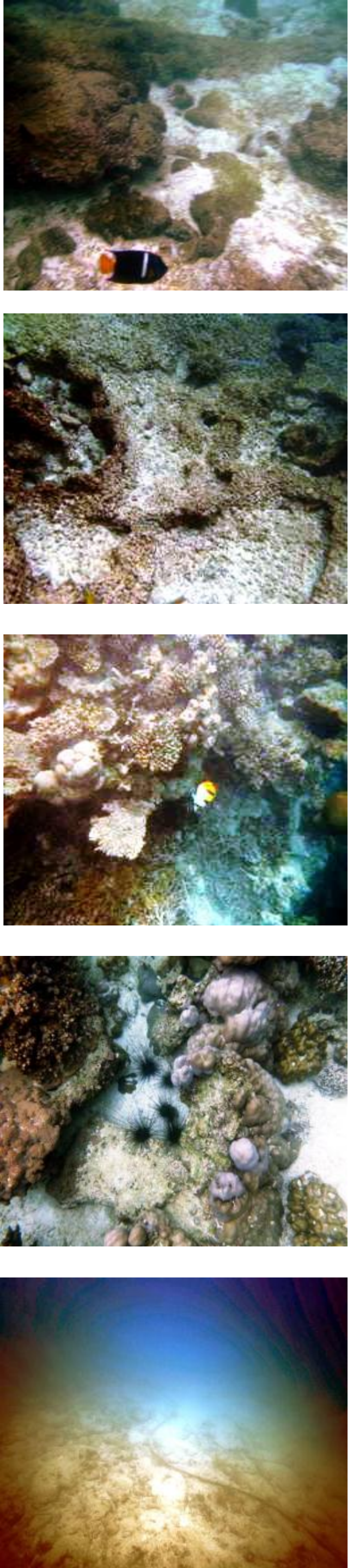}&
\includegraphics[width=.122\textwidth]{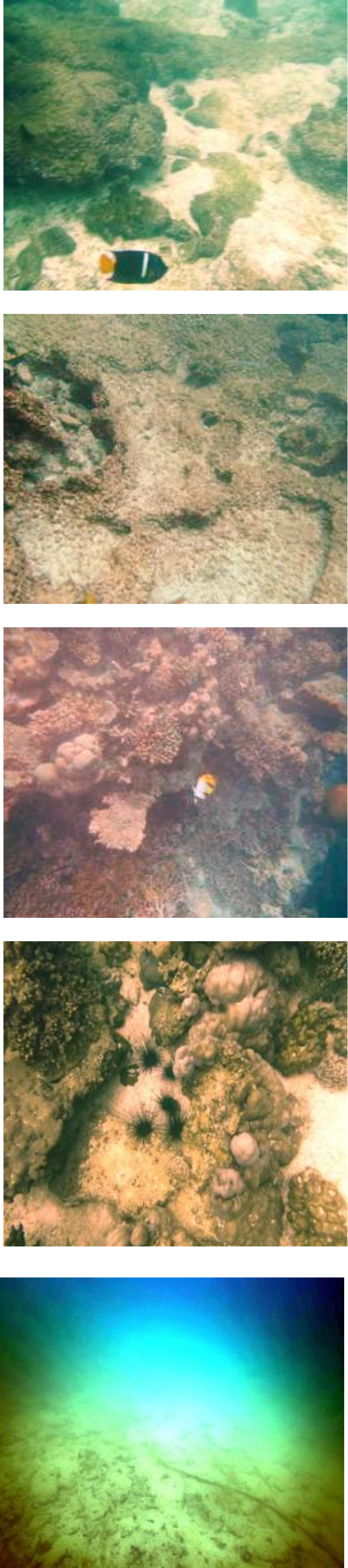}&
\includegraphics[width=.1205\textwidth]{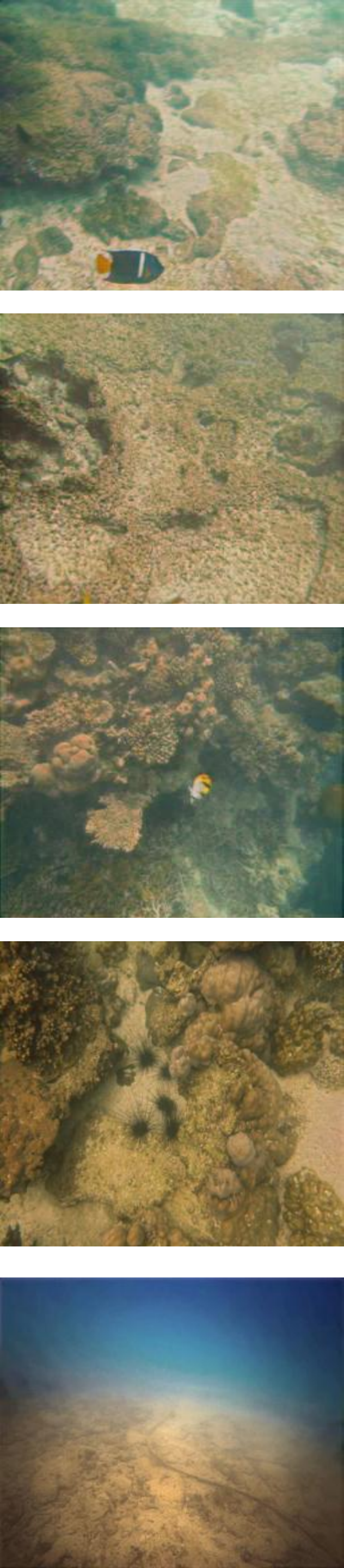}&
\includegraphics[width=.12\textwidth]{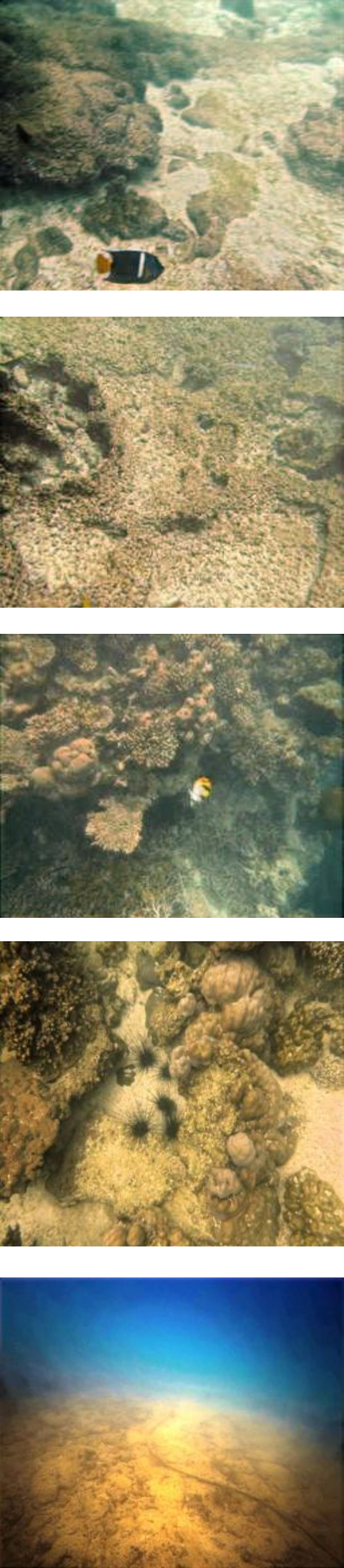}\\
(a) Real images& (b) RED &(c) UDCP&(d) ODM&(e) UIBLA&(f) UWCNN & (g) UWCNN+ \\
\end{tabular}
\caption{Results on real-world underwater images taken from the websites. Our method produces results without any visual artifacts, color deviations, and over-saturations. It also unveils spatial motifs and details.}
\label{fig:real_images}
\end{figure*}

 \begin{figure*}[!thb]
  \centering
\begin{minipage}[b]{0.3\linewidth}
  \centering
  \centerline{\includegraphics[width=4cm,height=3.2cm]{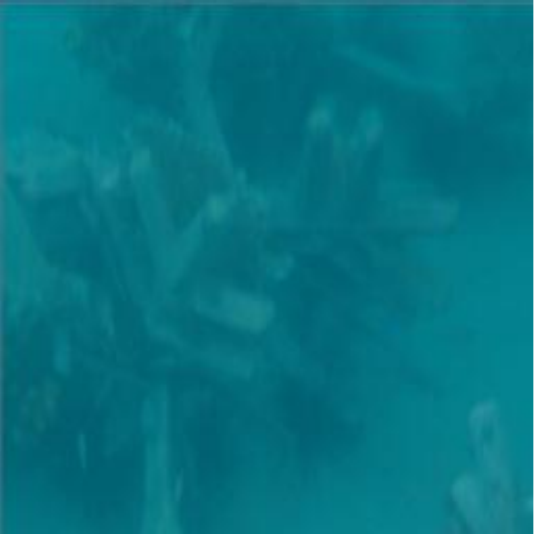}}
  \centerline{(a) Real image}\medskip
\end{minipage}
\begin{minipage}[b]{0.3\linewidth}
  \centering
  \centerline{\includegraphics[width=4cm,height=3.2cm]{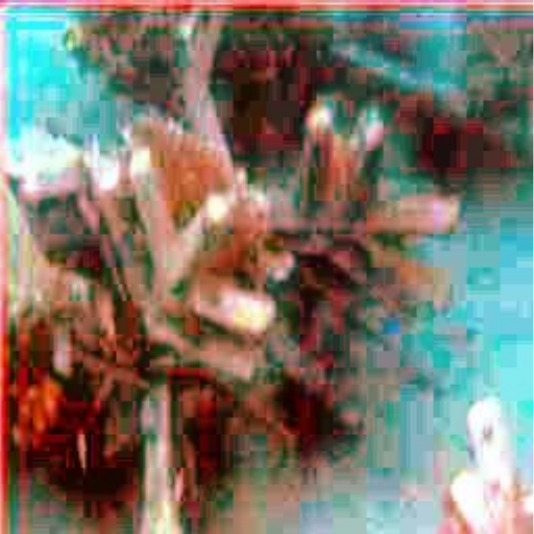}}
  \centerline{(b) ODM (incorrect reddish tones) }\medskip
\end{minipage}
\begin{minipage}[b]{0.3\linewidth}
  \centering
  \centerline{\includegraphics[width=4cm,height=3.2cm]{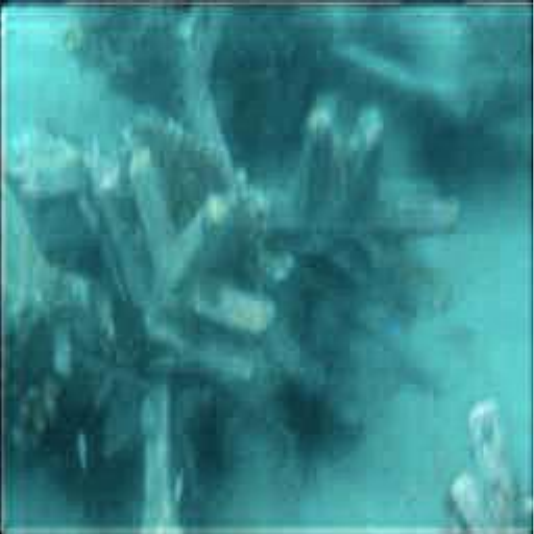}}
  \centerline{(c) UWCNN+}\medskip
\end{minipage}
 \caption{Comparison with ODM. (a) Real-world image. (b) Result produced by
ODM. It blindly introduces wrong colors, in particular, in red gamut. (c) Result produced by UWCNN+.}
\label{fig:failure_case1}
\end{figure*}

 \begin{figure*}[!thb]
  \centering
\begin{minipage}[b]{0.3\linewidth}
  \centering
  \centerline{\includegraphics[width=4cm,height=3.2cm]{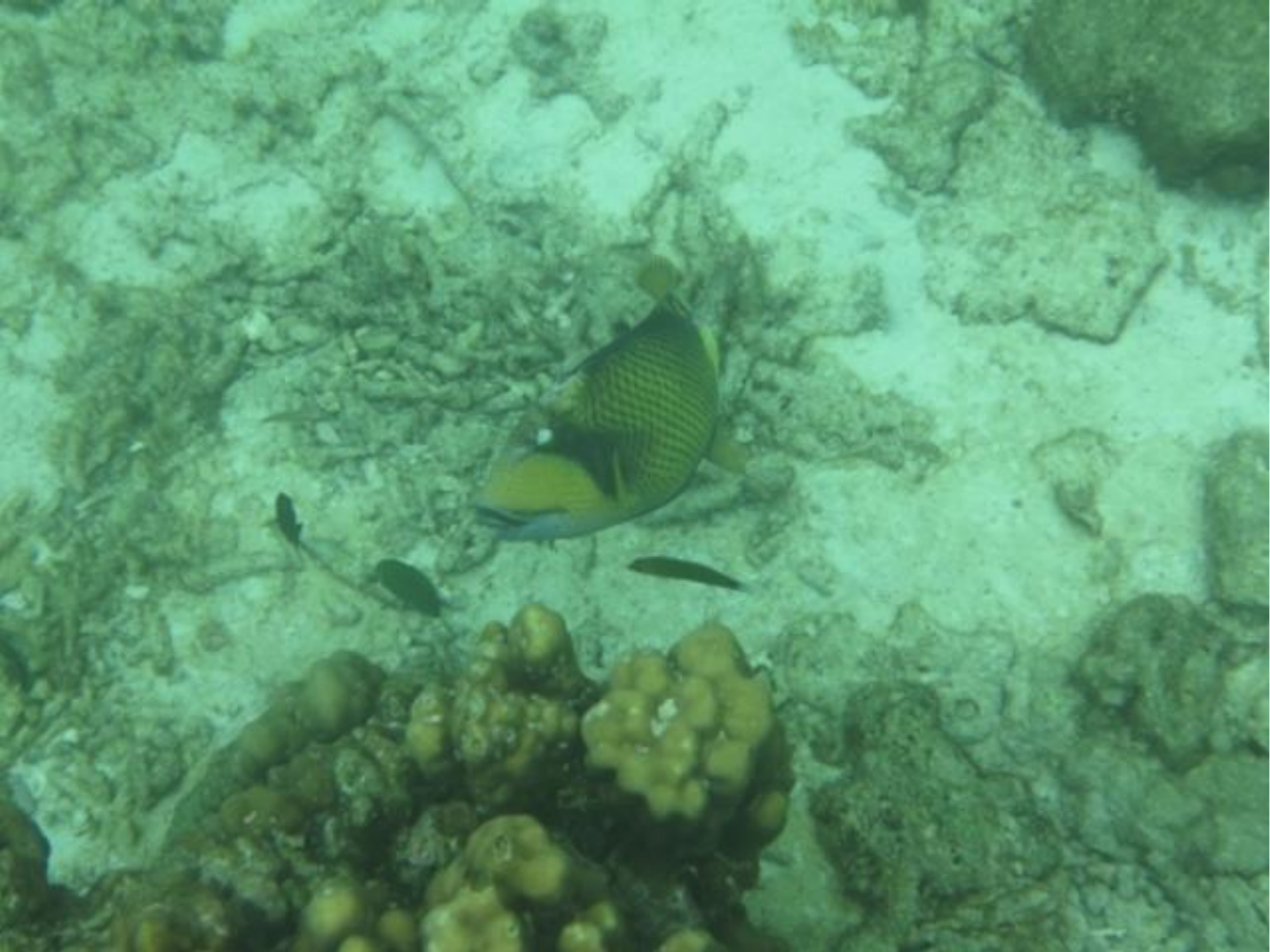}}
  \centerline{(a) Real image}\medskip
\end{minipage}
\begin{minipage}[b]{0.3\linewidth}
  \centering
  \centerline{\includegraphics[width=4cm,height=3.2cm]{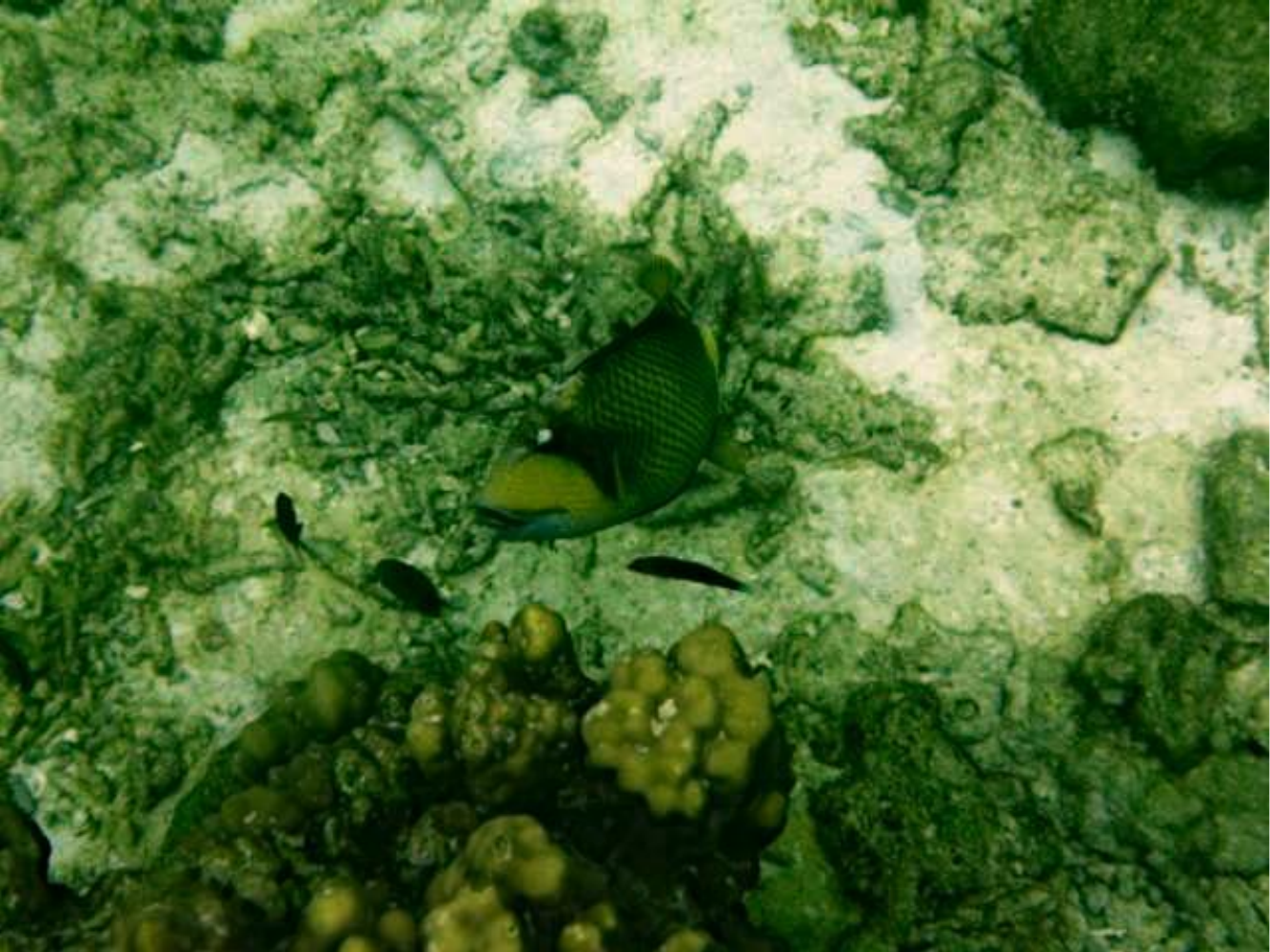}}
  \centerline{(b) UIBLA }\medskip
\end{minipage}
\begin{minipage}[b]{0.3\linewidth}
  \centering
  \centerline{\includegraphics[width=4cm,height=3.2cm]{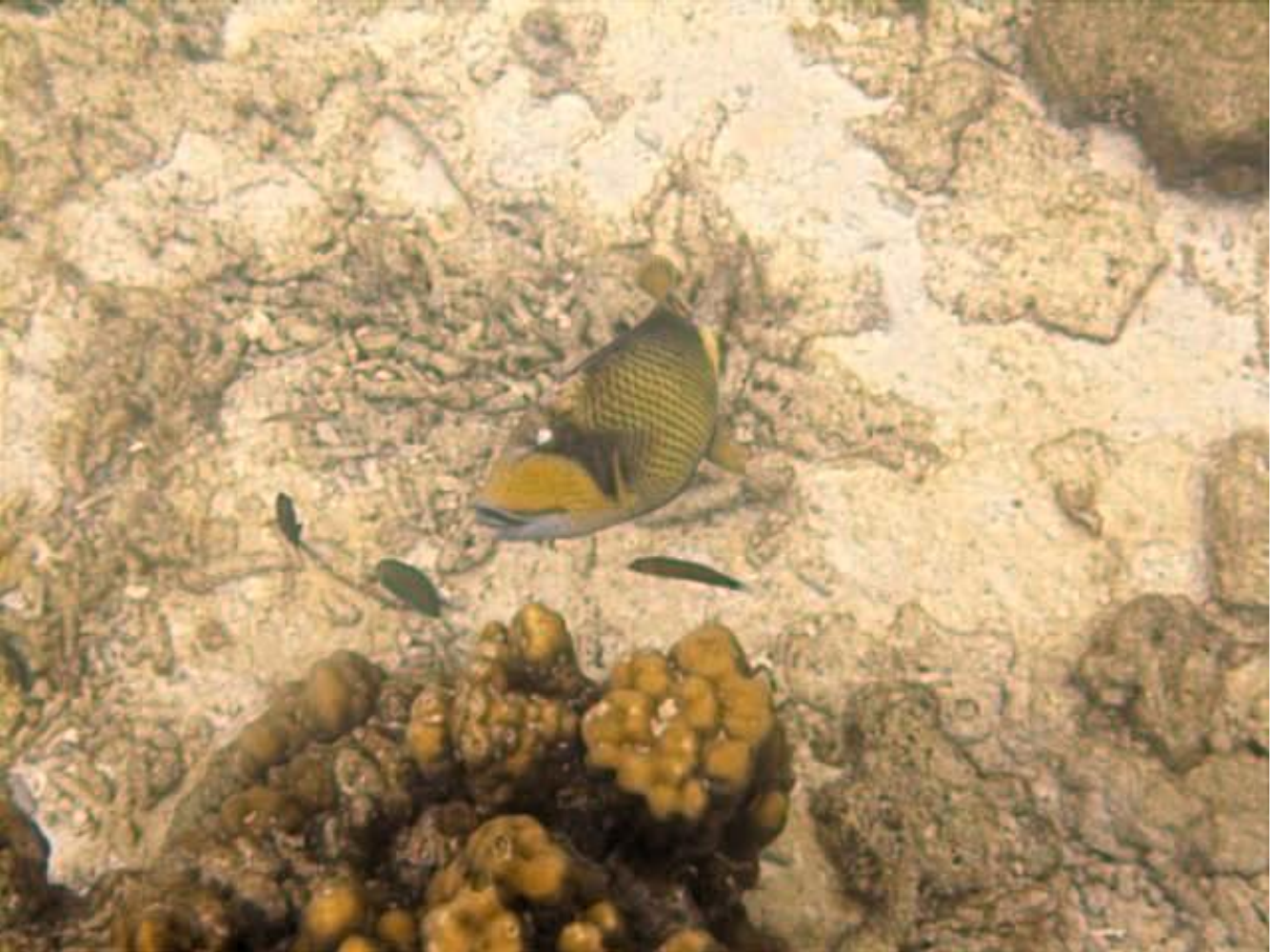}}
  \centerline{(c) UWCNN+}\medskip
\end{minipage}
 \caption{Comparison with UIBLA. (a) Real-world image. (b) Result produced by
UIBLA, which is a failure case since only greenish tones are enhanced. (c) Result produced by UWCNN+.}
\label{fig:failure_case2}
\end{figure*}

\subsection{Synthetic Underwater Images}

We first evaluate the capacity of our method in terms of quantitative metrics, thus we exploit synthetically generated underwater images, which is common in other computer vision tasks as they show the capacity of an algorithm in terms of quantitative metrics. We compare the UWCNN model with several state-of-the-art methods. In Figure~\ref{fig:synthetic_results}, we present the results of underwater image enhancement on the synthetic underwater images from our validation set.

As shown in Figure~\ref{fig:synthetic_results}(a), the synthetic underwater images accord with the measurement of \cite{Jaffe1976}. The RED method of \cite{Galdran142015} is effective for the clear types \ie, Type-1, Type-3, Type-5, and Type-I; however, for turbid one's \ie, Type-7, Type-9, Type-II, and Type-III, it leaves haze artifacts in those images, moreover, it introduces color deviations. Similarly, \cite{Drews152016} produces distinctly darkish results while ODM by \cite{Li32016} and UIBLA of \cite{Peng2017} improve the visibility in their respective outcomes while introducing artificial colors or color deviations. On the other hand, our UWCNN not only enhances the visibility of the images but also restores an aesthetically pleasing texture and vibrant yet genuine colors. In comparison to other methods, the visual quality of the UWCNN results resemble the ground-truth labels as shown in Figure~\ref{fig:synthetic_results}(g).

Furthermore, we quantify the accuracy of the recovered images on the synthetic validation set including 2495 samples for each type. In Table~II, the accuracy is measured by three different metrics: mean square error (MSE), peak signal to noise ratio (PSNR), and the structural similarity index metric (SSIM) \cite{Wang2004}. In case of MSE and PSNR metrics, the lower MSE (higher PSNR) denotes the result is more close to the label in terms of image content. In case of the SSIM metric, the higher SSIM scores mean the result is more similar to the label in terms of image structure and texture. Here, the presented results are the average scores. The values in bold represent the best results.

As visible, among all underwater image enhancement methods we tested, UWCNN model comes out as the best performer across all metrics and all degradation types, demonstrating its effectiveness and robustness. Regarding the SSIM response, our method is at least $10\%$ better than the second-best performer. Similarly, our PSNR is higher (less erroneous as indicated by the MSE scores) than the compared methods. The results in Table~2 also agree with the subjective images of Figure~\ref{fig:synthetic_results}.

\subsection{Real-world Underwater Images}

We also evaluate the proposed method on real-world underwater images. Visual comparisons with competitive methods are presented in Figure~\ref{fig:real_images}. The real-world underwater images are gathered from the Internet because there is no public underwater image dataset available. These real-world underwater images have varying tone, light, and contrast.

A first glance at Figure~\ref{fig:real_images} may give the impression that the results of ODM and UIBLA might be sharper; however, a careful inspection reveals that the ODM method causes over-enhancement and over-saturation (besides color casts) because the histogram distribution prior used in the ODM is not always valid. Similarly, the images produced by the UIBLA are not natural and consist of over-enhancement, a shortcoming of this method as the robustness value of the background light and the medium transmission score estimated by the prior are suboptimum. Figure~\ref{fig:failure_case1} and Figure~\ref{fig:failure_case2}  show the failure cases of the ODM and UIBLE methods. The methods of RED and UDCP have little effect on the inputs. In contrast, our UWCNN+ shows promising results on real-world images, without introducing any artificial colors, color casts, over- or under-enhanced areas. We do agree that our method does not enhance real-world images as accurately as it does the synthetic ones, which can be improved by having more underwater images in model reconstruction.

Observing the failure cases in Figure~\ref{fig:failure_case1} and Figure~\ref{fig:failure_case2}, one can find that the ODM method tends to introduce extra colors (\eg, the reddish color around the coral in Figure~\ref{fig:failure_case1}) while our approach improves the contrast,  similar performance to the ODM, but maintains a genuine color distribution of the original underwater image. For the failure case of the UIBLA method in Figure~\ref{fig:failure_case2}, it aggravates the greenish color and produces visually unpleasing results. In contrast, our method removes the color casts and improves the contrast and brightness, which generates better visibility and a pleasant perception.

\begin{figure}[!thb]
  \centering
\begin{minipage}[b]{0.3\linewidth}
  \centering
  \centerline{\includegraphics[width=8cm,height=7cm]{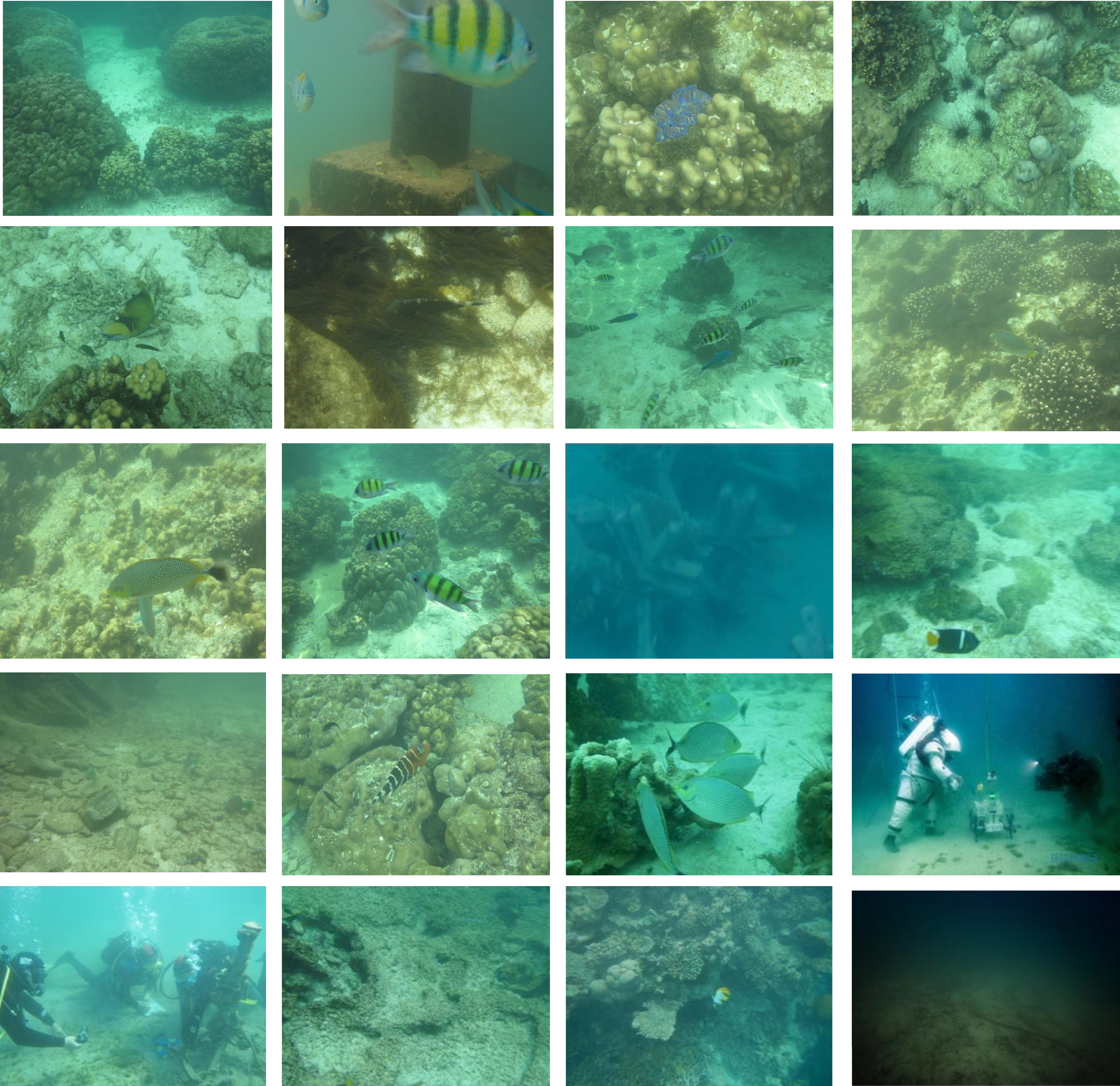}}
\end{minipage}
 \caption{Real-world underwater images with varying tone and degradation.}
\label{fig:dataset}
\end{figure}

We note that the assessments in \cite{Yang2015, Panetta2016} are slanted toward over-exposure or over-enhancement, where the histogram equalization method is regarded to yield better scores. For a more objective assessment, in addition to evaluating underwater image quality with several metrics, we conduct a user study to provide realistic feedback and quantify the subjective visual quality. We collect 20 real-world underwater images from the Internet and related papers. We show samples from this dataset in Figure~\ref{fig:dataset}. Some corresponding results are presented in Figure~\ref{fig:real_images}.

For the user study, we apply all methods to generate enhanced underwater images, randomize the order the results, and then display them on a screen to human subjects. There were 20 participants with image processing and computer vision expertise. Each subject ranks the results based on the perceived visual quality from 1 to 5 where 1 is the worst and 5 is the best. One expects that the results with high contrast, good visibility, natural color and authentic texture should receive higher ranks while the results with over-enhancement/exposure, under-enhancement/exposure, color casts, and artifacts should have lower ranks. The average subjective scores are given in Table~III. Our UWCNN+ receives the highest rankings, which indicates that our method can produce better performance on real-world underwater images from a subjective visual perspective.

\begin{table}[!t]
\renewcommand{\arraystretch}{1}
\caption{ User study on real-world underwater image dataset. The best result is in bold.}
\centering
\begin{tabular}{clcccccc}
  \hline
 \textbf  & \textbf{RED} & \textbf{UDCP} & \textbf{ODM} & \textbf{UIBLA} & \textbf{UWCNN+}\\
 \hline
Scores  & 2.95 & 2.55& 3.25 & 3.20  & \textbf{3.35} \\
\hline
\label{label:user study}
\end{tabular}
\vspace{-3mm}
\end{table}

\subsection{Ablation Study}
\label{sec:ablation}

To demonstrate the effect of each component in our network, we carry out an ablation study involving the following experiments:
\begin{enumerate}
\item UWCNN without residual learning (UWCNN-woRL),
\item UWCNN without dense concatenation (UWCNN-woDC),
\item UWCNN without SSIM loss (UWCNN-woSSIM).
\end{enumerate}
The quantitative evaluation is only performed on Type-1 and Type-III synthetic underwater image test sets due to the limited space. The average results in terms of MSE, PSNR, and SSIM are reported in Table \ref{table:ablation}. For MSE, a lower score is better. For PSNR and SSIM, the higher scores are better.

\begin{table}[!hp]

\begin{center}
\caption{Results for the Type-1 and the Type-III test sets. The best result for each evaluation is in bold, whereas the second best one is underlined.}
\begin{tabular}{ccccccc} \hline
 & \textbf{Types}    & \textbf{-woRL} & \textbf{-woDC}   &\textbf{-woSSIM} & \textbf{UWCNN}  \\ \hline
       &1     & 756.96 	& 648.18    & \textbf{398.77}    &\underline{587.70} \\
  MSE  &III   & 542.68 	& 789.76    & \textbf{402.92} 	 &\underline{456.40} \\ \hline
       &1     & 20.290 	& 20.805    & \textbf{22.902} 	 &\underline{21.790}  \\
  PSNR &III   & 21.556 	& 20.289    & \textbf{23.026} 	 &\underline{22.633}  \\ \hline
       &1     & \underline{0.8450}  & 0.8449  & 0.8214   &\textbf{0.8558}  \\
  SSIM &III   & \underline{0.8579}  & 0.8359  & 0.8151   &\textbf{0.8795} \\ \hline
  \label{table:ablation}
\end{tabular}
\end{center}
\end{table}

From Table \ref{table:ablation}, we see that replacing conventional learning strategy (UWCNN-woRL) with residual learning (UWCNN) could boost the performance. Comparing UWCNN with UWCNN-woDC, we observe that the dense concatenation also could improve the performance.

\begin{figure}[!h]
  \centering
\begin{minipage}[b]{0.47\linewidth}
  \centering
  \centerline{\includegraphics[width=4cm,height=3cm]{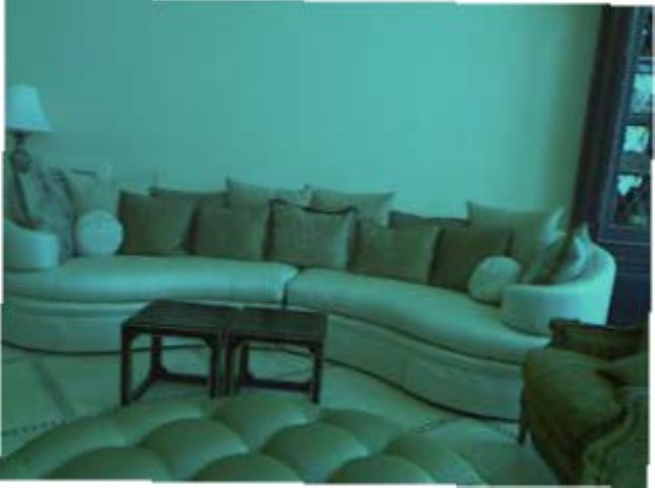}}
  \centerline{(a) Type-1 }\medskip
\end{minipage}
\begin{minipage}[b]{0.47\linewidth}
  \centering
  \centerline{\includegraphics[width=4cm,height=3cm]{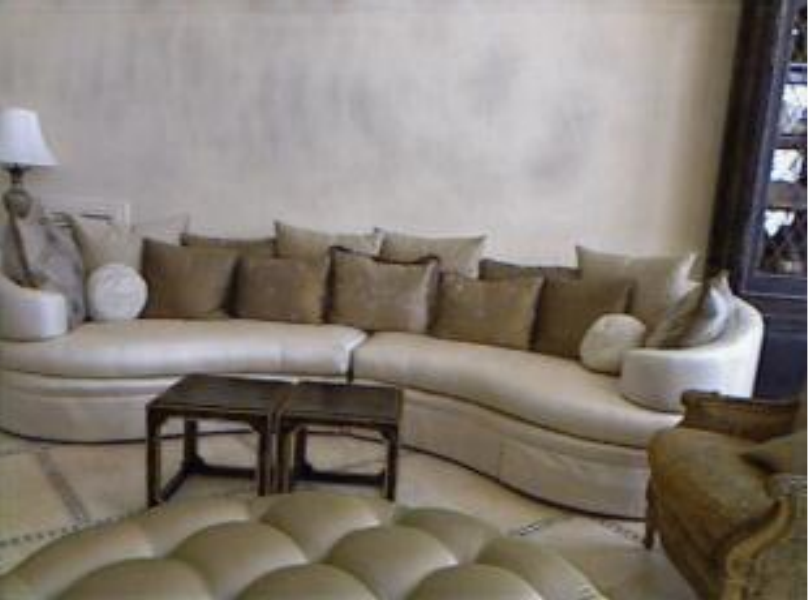}}
  \centerline{(b) -woSSIM }\medskip
\end{minipage}

\begin{minipage}[b]{0.47\linewidth}
  \centering
  \centerline{\includegraphics[width=4cm,height=3cm]{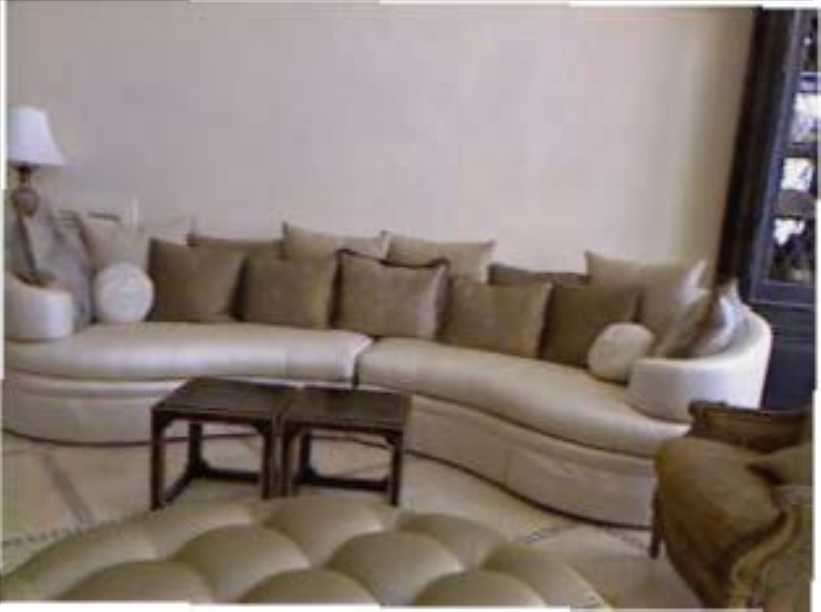}}
  \centerline{(c) UWCNN}\medskip
\end{minipage}
\begin{minipage}[b]{0.47\linewidth}
  \centering
  \centerline{\includegraphics[width=4cm,height=3cm]{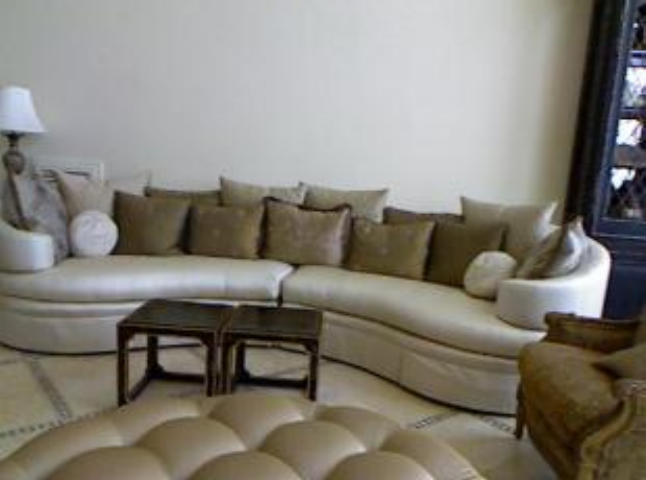}}
  \centerline{(d) GT}\medskip
\end{minipage}
 \caption{An example of the importance of the incorporation of SSIM loss. (a) The underwater image. (b) The result produced by UWCNN-woSSIM, which is a failure case since the background is not similar to the GT. (c) The result produced by our UWCNN. (d) The ground truth.}
\label{fig:SSIM_example}
\end{figure}

The use of SSIM loss (UWCNN) improves the structure and texture similarity at the cost of the decreased performance of MSE and PSNR (UWCNN-woSSIM). However, such a sacrifice is necessary for better subjective perception. Such an example is presented in Figure~\ref{fig:SSIM_example}, which demonstrates the importance of SSIM loss. In Figure~\ref{fig:SSIM_example}, after adding SSIM loss, the result of UWCNN has a more smooth background than that of UWCNN-woSSIM.

\section{Conclusion}

We presented convolutional neural network based underwater image enhancement network. We employ residual learning technique to train our models. Experiments are performed on synthetic and real-world images, which indicates the robust and effective performance of our method. To our advantage, our system only contains ten convolutional layers and 16 feature maps at each convolutional layer, which provides fast and efficient training and testing on GPU platforms.

To demonstrate the effectiveness of each component in our UWCNN network, we have carried out an ablation study. The results demonstrate that the residual learning, dense concatenation and SSIM loss used in our network boost the performance quantitatively and qualitatively.

In future, we plan to extend our current model to enhance hazy images and jointly produce image depth from a single hazy image. We will investigate using only one single model to predict the correct output from one single blind model of UWCNN to attain further accelerating in the process of UWCNN model enhancement.


\bibliographystyle{IEEEtran.bst}

\begin{IEEEbiography}[{\includegraphics[width=1in,height=1.25in,clip,keepaspectratio]{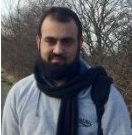}}]{Saeed Anwar} received Bachelor degree in Computer Systems Engineering with distinction from University of Engineering and Technology (UET), Pakistan, in July 2008, and Master degree in Erasmus Mundus Vision and Robotics (Vibot), jointly from Heriot watt University United Kingdom (HW), University of Girona Spain (UD) and University of Burgundy France in August 2010 with distinction. During his masters, he carried out his thesis at Toshiba Medical Visualization Systems Europe (TMVSE), Scotland. He has also been a visiting research fellow at Pal Robotics, Barcelona in 2011. Since 2014, he is a PhD student at the Australian National University (ANU) and Data61/CSIRO. He has also been working as a Lecturer and Assistant Professor at the National University of Computer and Emerging Sciences (NUCES), Pakistan. His major research interests are low-level vision, image enhancement, image restoration, computer vision, and optimization.
\end{IEEEbiography}

\begin{IEEEbiography}[{\includegraphics[width=1in,height=1.25in,clip,keepaspectratio]{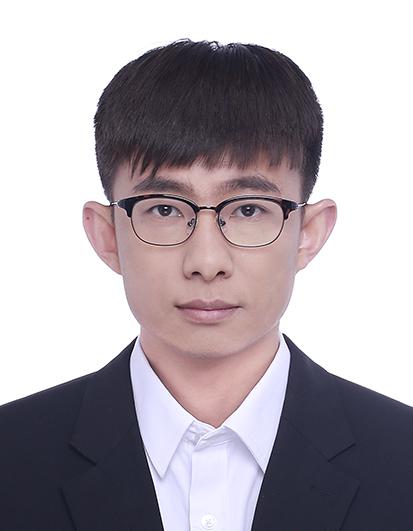}}]{Chong-yi Li}
received his Ph.D. degree with the School of Electrical and Information Engineering, Tianjin University, Tianjin, China in June 2018. From 2016 to 2017, he took one year study at the Research School of Engineering, Australian National University (ANU) as a visiting Ph.D. student supported by the CSC. Now, he is a lecture in the College of Electronic Information and Automation, Civil Aviation University of China (CAUC). His current research focuses on image processing, computer vision, and deep learning, particularly in the domains of image restoration and enhancement, such as images captured under the bad weather (hazy, foggy, sandy, dusty, rainy, snowy day) and special circumstances (underwater, weak illumination). He also focuses on other low-level vision problems, such as image/depth super-resolution reconstruction, image deblurring, image denoising, and multi-exposure image fusion.
\end{IEEEbiography}

\begin{IEEEbiography}[{\includegraphics[width=1in,height=1.25in,clip,keepaspectratio]{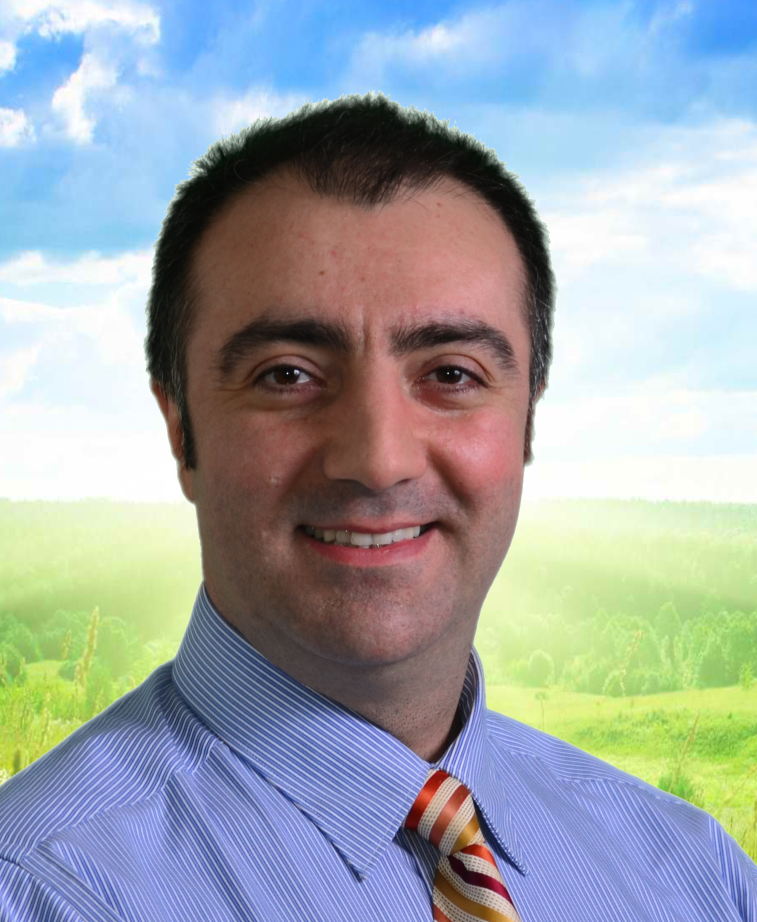}}]{Fatih Porikli}
is an IEEE Fellow and a Professor in the Research School of Engineering, Australian National University (ANU). He is acting as a Chief Scientist at Huawei. Previously, he led the Computer Vision Research Group at NICTA, and served as a Distinguished Research Scientist at Mitsubishi Electric Research Laboratories. He has received his PhD from New York University in 2002. Prof. Porikli is the recipient of the R\&D 100 Scientist of the Year Award in 2006. He won 5 best paper awards at premier IEEE conferences and received 5 other professional prizes. He authored more than 200 publications and invented 73 patents. He is the co-editor of 2 books. He is serving as the Associate Editor of several journals. His research interests include computer vision, deep learning, manifold learning, online learning, and image enhancement with commercial applications in autonomous vehicles, video surveillance, consumer electronics, industrial automation, satellite and medical systems.
\end{IEEEbiography}
 
\flushbottom

\end{document}